\documentclass{ieeeaccess}
\usepackage{cite}
\usepackage{amsmath,amssymb,amsfonts}
\usepackage{url}
\usepackage{graphicx}
\usepackage{textcomp}
\usepackage{xcolor}
\usepackage{caption}
\usepackage{subcaption}
\usepackage{algpseudocode}
\usepackage[ruled,vlined]{algorithm2e}
\usepackage{float}
\usepackage{afterpage} 
\usepackage{algorithm2e}
\usepackage{multicol}
\usepackage{setspace}
\usepackage{times}

\usepackage{bm}
\makeatletter
\AtBeginDocument{\DeclareMathVersion{bold}
\SetSymbolFont{operators}{bold}{T1}{times}{b}{n}
\SetSymbolFont{NewLetters}{bold}{T1}{times}{b}{it}
\SetMathAlphabet{\mathrm}{bold}{T1}{times}{b}{n}
\SetMathAlphabet{\mathit}{bold}{T1}{times}{b}{it}
\SetMathAlphabet{\mathbf}{bold}{T1}{times}{b}{n}
\SetMathAlphabet{\mathtt}{bold}{OT1}{pcr}{b}{n}
\SetSymbolFont{symbols}{bold}{OMS}{cmsy}{b}{n}
\renewcommand\boldmath{\@nomath\boldmath\mathversion{bold}}}
\makeatother

\def\BibTeX{{\rm B\kern-.05em{\sc i\kern-.025em b}\kern-.08em
    T\kern-.1667em\lower.7ex\hbox{E}\kern-.125emX}}

\begin{document}
\history{Date of publication xxxx 00, 0000, date of current version xxxx 00, 0000.}
\doi{10.1109/ACCESS.2024.0429000}

\title{Harmonic Oscillator based Particle Swarm Optimization}
\author{\uppercase{Yury chernyak}\authorrefmark{1},
\uppercase{Ijaz Ahamed Mohammad}\authorrefmark{1}, Nikolas Masnicak\authorrefmark{1}, Matej Pivoluska\authorrefmark{1,2} and Martin Plesch\authorrefmark{1,3}
}

\address[1]{Institute of Physics, Slovak Academy of Sciences,  Dúbravská cesta 5807/9, 845 11 Karlova Ves, Bratislava, Slovakia}
\address[2]{QTlabs,  Clemens-Holzmeister-Straße 6/6 Etage 6, 1100 Wien, Austria }
\address[3]{Matej Bel University, Národná ulica 12, 974 01 Banská Bystrica, Slovakia}
\tfootnote{We acknowledge the support of VEGA project 2/0055/23 and projects 09I03-03-V04-00425 and 09I03-03-V04-00685 of the Research and Inovation Authority.}

\markboth
{Author \headeretal: Preparation of Papers for IEEE TRANSACTIONS and JOURNALS}
{Author \headeretal: Preparation of Papers for IEEE TRANSACTIONS and JOURNALS}

\corresp{Corresponding author: Ijaz Ahamed Mohammad (e-mail: fyziijaz@savba.sk).}

\begin{abstract}
Numerical optimization techniques are widely used in a broad area of science and technology, from finding the minimal energy of systems in Physics or Chemistry to finding optimal routes in logistics or optimal strategies for high speed trading. In general, a set of parameters (parameter space) is tuned to find the lowest value of a function depending on these parameters (cost function).
In most cases the parameter space is too big to be completely searched and the most efficient techniques combine stochastic elements (randomness included in the starting setting and decision making during the optimization process) with well designed deterministic process. Thus there is nothing like a universal best optimization method; rather than that, different methods and their settings are more or less efficient in different contexts.
Here we present a method that integrates Particle Swarm Optimization (PSO), a highly effective and successful algorithm inspired by the collective behavior of a flock of birds searching for food, with the principles of Harmonic Oscillators. This physics-based approach introduces the concept of energy, enabling a smoother and a more controlled convergence throughout the optimization process.
We test our method on a standard set of test functions and show that in most cases it can outperform its natural competitors including the original PSO as well as the broadly used COBYLA and Differential Evolution optimization methods.   
\end{abstract}


\titlepgskip=-21pt

\maketitle

\section{Introduction}
\label{sec:introduction}

Meta-heuristic optimization techniques are a popular way to perform unconstrained minimization of complicated functions. 
These techniques are often inspired by natural phenomena, animal behaviors, or evolutionary concepts, making them easy 
to learn, implement, and hybridize. In addition, they are flexible and can be applied to various problems without altering their structure, 
treating problems as black boxes where only inputs and outputs matter. Unlike gradient-based approaches, meta-heuristics 
optimize stochastically without needing derivatives, making them suitable for complex problems for which derivatives are hard to obtain. 
Finally, their stochastic nature also helps avoid local optima, making them effective for challenging real-world problems with complex search spaces.

Despite quite a large number of different meta-heuristic methods published over the years, there is still an ongoing research in this area  
with current approaches being enhanced and new meta-heuristics being proposed frequently.
This is due to the No Free Lunch (NFL) theorem \cite{Wolpert1997}, which states that no single meta-heuristic is best for all optimization problems. 
An algorithm might perform well on one set of problems but poorly on another. This drives ongoing improvements and the development of 
new meta-heuristics, motivating our efforts to create a new one.

Meta-heuristics can be classified into single-solution-based and population-based methods. Single-solution methods, like Simulated Annealing \cite{Kirkpatrick1983}, start with one candidate solution that is improved iteratively. In contrast, population-based methods, like Particle Swarm Optimization \cite{Kennedy1995} (PSO), begin with multiple solutions that are enhanced over iterations. Advantages of population-based methods include information sharing among solutions, which leads to sudden jumps to promising areas, mutual assistance in avoiding local optima, and generally greater exploration compared to single-solution algorithms.

Population based meta-heuristics algorithms can further be classified into three main branches:
evolutionary, physics-based, and swarm intelligence algorithms.
Evolutionary algorithms, inspired by natural evolution, optimize by evolving an initial population of random solutions. The most popular algorithm in this class is the Genetic Algorithm (GA) \cite{Holland1992}, simulating Darwinian concepts. Each new population is formed by combining and mutating individuals from the previous generation, with the best individuals more likely to contribute, ensuring gradual improvement. Other evolutionary algorithms include Differential Evolution (DE)\cite{Storn1997}, Evolutionary Programming (EP)\cite{Yao1999}, Evolution Strategy (ES)\cite{Hansen2003}, Genetic Programming (GP)\cite{Koza1994}, and Biogeography-Based Optimizer (BBO)\cite{Ma2017}.

The second main branch of meta-heuristics is physics-based techniques, which mimic physical rules. Popular algorithms include Gravitational Local Search Algorithm (GLSA) \cite{DBLP:conf/ike/WebsterB03}, Big-Bang Big-Crunch (BBBC) \cite{EROL2006106}, Gravitational Search Algorithm (GSA) \cite{RASHEDI20092232}, Charged System Search (CSS) \cite{kaveh2010novel}, Central Force Optimization (CFO) \cite{formato2007central}, Artificial Chemical Reaction Optimization Algorithm (ACROA) \cite{ALATAS201113170}, Black Hole (BH) algorithm \cite{HATAMLOU2013175}, Ray Optimization (RO) algorithm \cite{KAVEH2012283}, Small-World Optimization Algorithm (SWOA) \cite{10.1007/11881223_33}, Galaxy-based Search Algorithm (GbSA) \cite{doi:10.1504/IJCSE.2011.041221}, and Curved Space Optimization (CSO) \cite{moghaddam2012curved}. These algorithms use a random set of search agents that move and communicate according to physical rules, such as gravitational force, ray casting, electromagnetic force, and inertia.

The third subclass of meta-heuristics is Swarm Intelligence (SI) methods, which mimic the social behavior of groups in nature. 
Similar to physics-based algorithms, these use search agents navigating through collective intelligence. 
The most popular SI technique is Particle Swarm Optimization (PSO), proposed by Kennedy and Eberhart \cite{Kennedy1995}, inspired by bird flocking behavior. PSO employs multiple particles that move based on their own best positions and the best position found by the swarm. Other algorithms in this class are  Ant Colony Optimization (ACO) \cite{Dorigo2004},  Artificial Bee Colony (ABC) \cite{Karaboga2007}, Bat-inspired Algorithm (BA) \cite{Yang2010}, Marriage in Honey Bees Optimization Algorithm (MHBO) \cite{934391}, Artificial Fish-Swarm Algorithm (AFSA) \cite{li2003new}, Termite Algorithm (TA) \cite{Martin2006}, Wasp Swarm Algorithm (WSA) \cite{10.1007/978-3-540-71618-1_39}, Monkey Search (MS) \cite{10.1063/1.2817338}, Bee Collecting Pollen Algorithm (BCPA) \cite{10.1007/978-3-540-85984-0_62},
 Cuckoo Search (CS) \cite{5393690},
 Dolphin Partner Optimization (DPO) \cite{5209016}, Firefly Algorithm (FA) \cite{yang2010firefly}, Bird Mating Optimizer (BMO) \cite{https://doi.org/10.1002/er.2915},
 Krill Herd (KH) \cite{GANDOMI20124831},
 Fruit fly Optimization Algorithm (FOA) \cite{PAN201269} and Grey Wolf Optimizer \cite{mirjalili2014grey}.


Among this pool of methods, Particle Swarm Optimization (PSO) stands out as a significant and relevant algorithm for several reasons such as its simplicity and ease of implementation, as well as outstanding results in test cases and real deployment. 
Unlike Genetic Algorithms (GA) and Ant Colony Optimization (ACO), PSO requires fewer parameters to adjust, resulting in a reduced computational burden \cite{Sivanandam2008}, \cite{en16031152}, \cite{nayak202325}. 
PSO also benefits from its internal "memory" capabilities; it leverages previously known best positions to enhance search efficiency, unlike GAs. 
Its scalability is another key feature, as PSO can effectively handle large-scale, high-dimensional complex optimization problems. In terms of global optimization, PSO demonstrates the ability to avoid multiple local minima, allowing it to navigate rugged landscapes and identify the global minimum. 
Additionally, PSO's flexibility enables it to be easily hybridized with other optimization techniques, further improving its results \cite{WANG2018162},\cite{THANGARAJ20115208},\cite{qiao2024hybrid}.

As one could expect, despite the numerous benefits mentioned above, PSO has a few drawbacks to show as well. One of the major issues it faces is the occasional uncontrolled movement of some of the birds in the flock, which can lead to poor convergence. More specifically, even if the optimal points determined by the birds so far are confined to a small region, the birds can gain very high velocities, resulting in an expansion of the search space into irrelevant, large regions. On other occasions, birds lose their velocity very quickly and then proceed to get stuck on a point despite the optimization needing to continue.

A natural approach to address these issues is to get under control the reduction as well as the possible increase of the velocity of the birds. As we show in the next section, however, this is not that easy, and for this purpose, the commonly used settings for the hyper-parameters of PSO are not always practical. This is why we introduce, in Section III, a new population-based meta-heuristic method, that introduces the concept of energy (a combination of velocity and distance from the optimal positions known thus far) into PSO. This allows to control the convergence of the method independently of other parameters, while still keeping all the existing PSO features intact. While the main design inspiration is based on the PSO algorithm, this method is better classified as a physics based meta-heuristic, because the movement of the particle population is governed by the physics of harmonic oscillators -- hence we name the method as Harmonic Oscilator based Particle Swarm Optimization (HOPSO). 

Results of the HOPSO method are described in Section IV, showcasing its capabilities on $12$ different test functions, while the Section V concludes our findings.

\section{Introduction to PSO and its problems}


\quad The Particle Swarm Optimization (PSO) is a meta-heuristic population based optimization method that works by simulating the social behavior of a flock of birds or a school of fish. 
In these social systems, the movement of the swarm species was observed to be a form of optimization in their search for food. 

Modeling after this swarm behavior, the PSO was developed such that each particle represents a potential solution, and it modifies its position in the search space according to the individual experience (a “cognitive” term) and the group experience (a “social” term). 

PSO is simple and easily implementable, and it can ensure fast convergence to a satisfactory solution with a small number of control parameters. It works well even when the search space is large or when the problem is highly non-linear or multi-modal.

PSO also frequently outperforms other algorithms on problems with a smooth landscape, where the optimization process would benefit from a more exploratory approach that can be realized by the collective behavior of the swarm -- this is particularly important in the newly developing field of Variational Quantum  Algorithms \cite{nano12020243}, \cite{PhysRevA.107.032407}. 

To briefly describe the working of PSO, the algorithm begins with a population of $N$ particles, each representing a candidate solution in the \textit{d}-dimensional search space of the optimization problem under consideration.
The positions of these particles are randomly initialized within predefined search space boundaries and move according to specific update equations in discrete time steps, i.e. iterations.
Specifically, the determination of the position in the subsequent iteration is dependent on the current position with an additional velocity-vector that drives the particle to a new, and ideally better, position in the search space.
This velocity vector for the next iteration incorporates three crucial components: the inertia (previous velocity), the cognitive component, and the social component.
The cognitive component represents the element-wise difference between the personal best position vector and its current position vector, while the social component represents the element-wise difference between the global (swarm) best position vector and the particle's current position.
Each of these is then scaled by the `cognitive' and `social' coefficients, denoted as $c_1$ and $c_2$, respectively.
These coefficients represent the relative importance assigned to the particle's personal best position (stored in its individual memory) and the swarm's global best position when calculating the velocity for the next iteration, thereby determining the relative importance of the particle's personal experience versus the swarm's collective knowledge in calculating the next velocity and position. 

\begin{figure}[h!]
    \centering
    \includegraphics[width=0.5\textwidth]{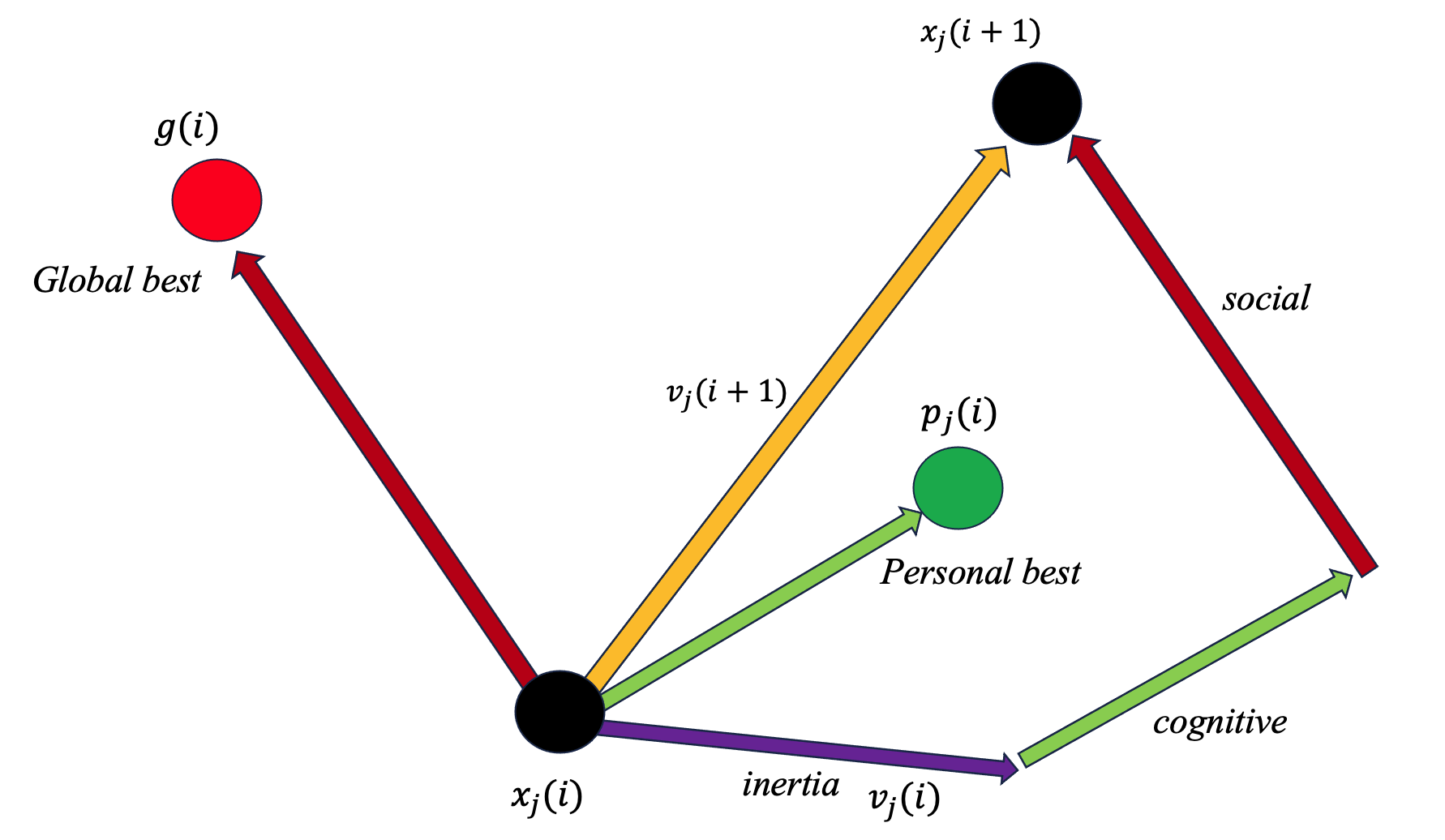}
    \caption{An example of the movement of a particle in a two-dimensional space based on the PSO algorithm in a single iteration. 
    An inertia term given by velocity that drives the particle in some direction (violet arrow), a memory term ($p_j$) that influences the particle's trajectory based on its best known position (green arrow), and a global-best cooperation term ($g$) that reflects the best result amongst the entire swarm (red) constitute the particle's projected movement (yellow arrow). The $i$ indicates iteration, while $j$ indicates particle number.}
    \label{fig:pso_figure}
\end{figure}

Moreover, each of these terms is adjusted by a unique random factor, thereby introducing stochasticity into the search process. This movement is illustrated in figure \ref{fig:pso_figure}. To allow convergence of the whole system, a damping factor in the form of a \textit{constrictor factor} $\chi$ was introduced in \cite{Kennedy2002}. The update equations of this variation of the PSO algorithm are introduced as follows with a description of its variables in Table \ref{tab:PSOParameters}. \\

\textbf{Velocity Update:}
\begin{equation}\label{vel_equation}
\begin{split}
v_{j,d}(i+1) = \chi(v_{j,d}(i) + c_1r_1(p_{j,d} - x_{j,d}(i)) \\
               + c_2r_2(g_d - x_{j,d}(i)))
\end{split} 
\end{equation}

\textbf{Position Update:}
\begin{equation}\label{pos_equation}
x_{j,d}(i + 1) = x_{j,d}(i) + v_{j,d}(i + 1).
\end{equation}

Balancing exploration capabilities and reasonable convergence is the main challenge across all PSO variants. A low constriction factor causes particles to stop moving too quickly, while high values (leading to low damping) cause particles to spread into large regions far from any optima. A specific form of the constrictor factor, $\chi$, has therefore been derived from the stability analysis of the PSO system to address these concerns regarding the values of the velocity term. Particularly, the velocity update equation can be viewed as a second-order difference equation which occurs when removing random variables from the PSO equation and is then examined using roots of its characteristic equation. The roots, $\lambda$, determine the behavior of the velocity over time and thereby, for the system to be stable (converge), the absolute value of the roots must be less than 1. Thus, the form of $\chi$ must be such that the eigenvalues of the system
are indeed less than 1. Also, if the parameters $c_1$ and $c_2$ are too large, the system also becomes unstable, leading to divergence of the particle velocities. Their sum $\varphi = c_1 + c_2$ directly impacts the constrictor factor $\chi$ and, consequently, the magnitude of the velocity update. Larger values of $\varphi$ causes the constrictor factor to dampen the velocities more significantly to prevent runaway velocities that lead to divergence, therefore allowing a better and more controlled convergence behavior. Lastly, the presence of the square root and absolute value in $\chi$ ensures that they reduce gradually as they approach an optimal solution, preventing overshooting or oscillatory behavior. From these observations in their analysis, Clerc and Kennedy proposed the mathematically designed form of $\chi$ to be

\begin{equation}\label{constriction_factor}
\chi = \frac{2}{\left| 2 - \varphi - \sqrt{\varphi^2 - 4 \varphi} \right|}
\end{equation}
\begin{equation}\label{varphi_no_rng}
\varphi = c_1 + c_2.
\end{equation}

Empirical studies suggest that commonly used parameters to ensure convergence are \(c_1 = c_2 = 2.05\)\cite{4223164}, with a constriction factor \(\chi = 0.7298\), derived from the analysis of PSO without randomness. 

\begin{table}[h!]
\caption{Explanation of PSO parameters}
\centering
\begin{tabular}{p{1.5cm}|p{5.5cm}}

\textbf{Parameter} & \textbf{Description} \\

$v_{j,d}(i)$ & Velocity of particle $j$ in dimension $d$ at iteration $i$ \\
$x_{j,d}(i)$ & Position of particle $j$ in dimension $d$ at iteration $i$ \\
$\chi$ & Constrictor factor (damping) \\
$c_1$ & Cognitive coefficient, attraction towards the particle's best known position \\
$c_2$ & Social coefficient, attraction towards the swarm's best known position \\
$r_1, r_2$ & Random values uniformly distributed in the range $[0,1]$ \\
$p_{j,d}$ & $d$-th dimension of the best known position of particle $j$ \\
$g_d$ & $d$-th dimension of the global best known position \\

\end{tabular}
\label{tab:PSOParameters}
\end{table}


Let us now analyse in detail the potential for velocity explosions in PSO, a phenomenon described by Kennedy and Eberhart \cite{Kennedy2002}. This occurs when particle velocities significantly exceed the characteristic scale of the search space during the optimization process, leading to swarm divergence, as these high-velocity particles continue to traverse the parameter space rapidly and their ability to effectively locate optima of the cost function becomes severely compromised. This behavior thereby undermines the balance between exploration and exploitation that is crucial for the PSO's effectiveness.

In order to understand why these velocity explosions occur, let us briefly analyse the time evolution of a single bird in our swarm. For simplicity, let us consider only one dimension, as each of the dimensions are behaving independently until a new optimum is identified. In this case, the position update equation \eqref{pos_equation} and velocity update equation \eqref{vel_equation} can be represented with a matrix \textit{M} acting on the vector with position and velocity coordinates which thereby provides a compact way to describe the coupled dynamics of both velocity and position written as presented in 
\eqref{vecotr_evolution}

\begin{equation}\label{vecotr_evolution}
    \Vec{P}_{t+1} = M\Vec{P}_t, \quad \Vec{P}_t = (v_t,y_t),
\end{equation}
where 
\begin{equation}\label{position_shift}
    y_t = \dfrac{\varphi_1 g +\varphi_2 p}{\varphi_1 + \varphi_2} - x_t,
\end{equation}
with $\varphi_1 = r_1c_1$ and $\varphi_2 = c_2r_2$ and the dynamical matrix $M$ governing the time evolution defined as
\begin{equation}\label{dynamical_matrix}
  M=  \begin{pmatrix}
        \chi & \chi\varphi \\
        -\chi & 1 - \chi\varphi
    \end{pmatrix},
\end{equation}
where $\varphi = \varphi_1 + \varphi_2$. 

Notice that equation \eqref{vecotr_evolution} only represents a single iteration step but since the random numbers $r_1$ and $r_2$ are changing in each iteration, to represent the state of the particle after several iterations we need to accumulate the effects of all previous iterations, therefore the state vector $\Vec{P}_t$ of a bird that was initially at position $x_0$ with velocity $v_0$ can be described as a product of transformations

\begin{equation}\label{random_state_update}
    \Vec{P_t} = \prod_{i = 0}^{t} M_i\Vec{P_0}, \quad \Vec{P}_0 = (v_0,y_0).
\end{equation}

If no new local or global optimum is found, it is expected that our bird will gradually converge and its velocity will decay to zero. 
As analysed in \cite{Kennedy2002}, when one keeps the same value of $\chi$ and $\varphi$ throughout the entire simulation (removing the stochastic nature of the method), the sufficient condition for convergence is
\begin{equation}\label{convergence_condition}
    \max(|\lambda_1|,|\lambda_2|) < 1,
\end{equation}
where $\lambda_1$ and $\lambda_2$ are the eigenvalues of $M$ given by
\begin{equation}\label{eigenvalues}
    \lambda_{1,2} = \dfrac{1}{2}\Big(1 + (1-\varphi)\chi \pm \sqrt{((\varphi-1)\chi-1)^2 - 4\chi}\Big),
\end{equation}
as both the eigenvectors are multiplied by $\lambda_{1}^{t}$ and $\lambda_{2}^{t}$ which both converge to $0$.

This is no longer the case when one includes the randomness into PSO.  In this case the behavior is governed by a product of $t$ different dynamical matrices in \eqref{random_state_update}. Here, despite the fact that each individual matrix $M$ has the magnitude of both eigenvalues below $1$, it is in general not the case for their product, i.e. subsequent multiplication of matrices, with each of their eigenvalue being smaller than one, to also form a matrix with a very small eigenvalue, therefore having large eigenvalues will lead to explosions in velocities.

To analyse the product of different matrices, one has rather to look into the singular values of the dynamical matrix $M$ given by
\begin{multline}\label{singular_values}
\sigma^2_{1,2} = \frac{1}{2}\Big(2\chi^2\big((c_1r_1 + c_2r_2)^2 +1\big) - 2(c_1r_1 + c_2r_2)\chi + 1 \\
\pm \sqrt{(2\chi^2\big((c_1r_1 + c_2r_2))^2 +1) - 2(c_1r_1 + c_2r_2)\chi + 1)^2 -4\chi^2}\Big).
\end{multline}
Singular values express the limits of maximum possible prolongation or contraction of a vector that is multiplied by a given matrix. To have the length of the vector under control, one optimally needs to have both singular values being similar, and for smooth convergence it must be smaller than one but close to one. 

\begin{figure}[h!]
    \centering
    \includegraphics[width=0.5 \textwidth]{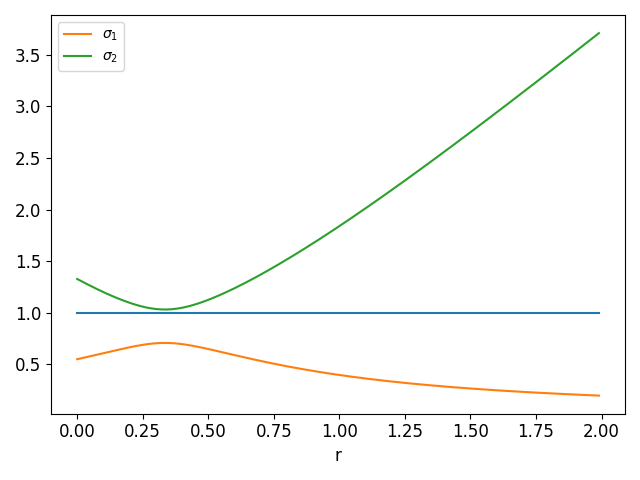}
    \caption{Singular values of dynamical matrix $M$ for $c_1 = c_2 = 2.05$, in which case $\chi = 0.729$. One can see that in the region of small sum of random numbers both singular values are close to one and their average is below one, so one can expect some kind of reasonably small convergence. However, for large sum of random numbers this is not the case -- one singular values almost diminishes and the other one reaches almost the value of $4$, thus one can, depending on the starting position and exact combination of random numbers obtain both very fast dying out, as well as velocity and position explosions. Let us note, however, that using two independent random numbers $r_1$ and $r_2$ makes the probability distribution of its sum higher, when the sum of $r_1$ and $r_2$ is around 1, making the explosions less probable.}
    \label{fig:sing_values}
\end{figure}

Using a simplification of $c_1 = c_2 = 2.05 = c$ along with definition of $r= r_1 + r_2$ as the sum of random numbers used in the specific step, expression (\ref{singular_values}) becomes 
\begin{multline}\label{singular_values}
\sigma^2_{1,2} = \frac{1}{2}\Big(2\chi^2\big((cr)^2 +1\big) - 2cr\chi + 1 \\
\pm \sqrt{\big(2\chi^2(cr)^2 +1) - 2cr\chi + 1\big)^2 -4\chi^2}\Big).
\end{multline}
Singular values for matrix (\ref{dynamical_matrix}) are shown in Fig. \ref{fig:sing_values} -- note that the sum of random numbers runs from $0$ to $2$.

Results of this theoretical analysis have been confirmed by having studied the behavior of the method in specific real-world cases. First of all, the explosions have been seen as resulting from a specific series of random numbers, where large and small numbers have regularly followed after each other. This is because the specific form of $M$ matrix causes rotation in the governing vector (large velocity in one step induces large distance in the next one), leading to the application of the second singular value in the next step. If, however, a large random number is applied when the larger singular value is active, along with a small random number combined with smaller singular value, the resulting behavior is divergent.

This understanding is also confirmed by the observed fact, that when one decided to use only a single random number (i.e. defining $r_1=r_2$), the velocity explosions happened regularly, while other birds died very quickly. This is due to the fact that the extreme singular values were achieved with a much higher probability compared to when there were two independent random numbers.  

It is clear that achieving a smooth and controlled convergence of the method with its existing governing equation, a very detailed control of randomness applied would be necessary. This would, however, compromise the stochastic nature of the method -- the more restrictions that are applied, the higher is the probability of not reaching the global maximum. Thus, we suggest to re-formulate the PSO idea in a more physical framework, introducing the concept of energy, its conservation and loss via damping, together with possible energy boosts connected with finding a new global or local optimum. 

Achieving smooth and controlled convergence with the current governing equations would require precise management of the randomness applied. However, this would undermine the stochastic nature of the method—the more constraints imposed, the higher the likelihood of failing to reach the global maximum. Therefore, we propose reformulating the PSO approach within a more physical framework, introducing concepts such as energy, conservation, and loss through damping, along with potential energy boosts associated with finding new global or local optima.

\section{Harmonic Oscillator based Particle Swarm Optimization -- HOPSO}

The general framework of energy consists of two basic types -- kinetic energy, defined by the movement of specific particles, and potential energy, defined in most cases by the position of particles and expressing the potential to gain (or loose) kinetic energy by changing its position. Potential energy can only be associated by conservative forces, such as gravity or electromagnetic forces, whereas friction forces are typical examples of non-conservative forces that lead to loss of energy (or, more precisely, to dissipation into heat). 


While there exist optimization approaches that are based on gravity \cite{RASHEDI20092232}, \cite{KHAN2021104263}, which is mathematically very similar to the model of charges, we consider it as not a very promising approach. Firstly, the energy diverges to negative infinity while reaching the center of gravity, which leads, again, to explosions of velocities.  Secondly, for long distances the attractive force only increases in a minor way, allowing the particles to fly very far from the attractor. 

For such reasons, we were inspired to use spring forces, e.g. forces that are linear with the distance between the attractor and the position of the particle, and can be associated with an ideal spring. This concept is optimal as for small distances the potential energy converges to $0$ while for large distances the energy grows quadratically, essentially bounding the particle to a well defined region. 

One challenge is that when a single spring is used across multiple dimensions, the total energy depends in a complex non-trivial way on the position in all dimensions. To simplify the process, we propose a model in which an independent virtual spring is assigned to each dimension, attracting the particle within that dimension only. This approach maintains the principle of energy conservation while enabling faster and simpler calculations. It also allows for independent adjustment of constraints in different dimensions, as the energy is now decoupled for each parameter.

Now the movement of particles, if modelled in continuous time, would reflect swinging on a spring with a center defined by an attractor (a suitable combination of social and cognitive term) independently in each single dimension. The period of these oscillations is defined by a combination of  a virtual mass of that particle and the stiffness of the spring -- each of which can be chosen arbitrarily. This movement has a simple analytical solution -- harmonic motion. 

In the optimization process, there is no need to model the entire motion. Instead, snapshots are taken at different moments during the harmonic oscillations. Randomness starts to influence the process in HOPSO at this stage. Rather than altering the potential (by adjusting the constants that link position and velocity changes) which would cause unpredictable energy fluctuations, we instead allow the particles to oscillate harmonically and observe them as snapshots at different time intervals.

In contrast to the original PSO where the damping influences multiple aspects including the social and cognitive terms through the constrictor factor, here the damping constant solely governs energy dissipation which directly relates to the searching ability of the particle. In the results section we will show how to use this parameter, indicating that not only does this parameter offer more tunability in governing convergence via this physically-inspired model, but also that this damping parameter is an easily controlled independent parameter (i.e. can be classified as a "free parameter"). This flexibility offers a significant advantage for optimization problems, and is one of the crucial advantages of the HOPSO algorithm over not only the standard PSO method, but other non-gradient methods as well, presented in our results section.

More formally, each particle's position in each dimension  is defined as the solution of a damped harmonic oscillator. 
The solution is given by the following equation:
\begin{equation} \label{eq3}
    x(t) = A_{0}e^{-\lambda t} \cos{(\omega t + \theta)} +x_0,
\end{equation}
where, $x(t)$, $A_{0}, \lambda, \omega$ and
$\theta$ represent the position of the particle at time $t$, initial amplitude, damping factor, angular frequency and initial phase of the oscillation, respectively, whereas the $x_0$ is the position of the attractor in that direction.

The velocity of the particle at any given time $t$ can be obtained by differentiating \eqref{eq3}
\begin{equation} \label{eq4}
    v(t) = -\omega (A_{0} e^{-\lambda t} \sin{(\omega t + \theta)}) -\lambda(x(t)).
\end{equation}

The time of measurement is chosen randomly within the interval [0,$t_{ul}$] for each particle and in each dimension, where $t_{ul}$ is the upper limit of the time sampling range and typically is chosen as the period of oscillation. 
The iterative change in parameter $t$ is defined as \eqref{eqt}
\begin{equation}\label{eqt}
    t_{i+1} = t_{i} + rand[0,t_{ul}],
\end{equation}
where the parameter $i$ represents the index of iteration.


The optimization begins by initializing the particles at random positions along with random velocities in the parameter landscape. 
The initial positions are noted as the initial personal-best positions. The global-best position is noted as the best position out of all the personal-best positions of all the particles. This global best position corresponds to the lowest value of the cost function within the swarm.

Each particle then oscillates about an attractor independently in each dimension. This attractor can be calculated as
\begin{equation}\label{eq5}
    a_{j,d} = \frac{c_{1}p_{j,d} + c_{2}g_{d}}{c_{1}+c_{2}},
\end{equation}
where $a, p, g$ represent, respectively, the position of the attractor, personal best position of the particle and the global best position for the $j^{th}$ particle in $d^{th}$ dimension. 
The $c_{1}$ and $c_{2}$ terms represent the weights of attraction towards the personal-best and global-best positions, respectively. 
Typically, the values $c_{1}$ and $c_{2}$ are set equal, so the attractor lies equidistant between the personal best position and the global best position. 

We solve \eqref{eq3}  and \eqref{eq4} to obtain the initial amplitudes $A_{0}$ for each particle and in each dimension by choosing the initial time as $t=0$:

\begin{equation} \label{eq6}
    A_{0} = \sqrt{(x(0)-a)^2+\frac{(v(0)+\lambda(x(0)-a))^2}{\omega^2}}.    
\end{equation}
Once the initial oscillation amplitude $A_{0}$ is determined, the initial phase of the oscillation $\theta$ can be calculated as
\begin{equation}\label{eq9}
    \theta = \arccos{\frac{x(0)-a_{j,d}}{A_{0}}}.
\end{equation}
We then let the particles oscillate in time, while this initial amplitude of oscillation $A_{0}$ decays as $A_{0}e^{-\lambda t}$.  For every iteration, we stop the clock at a random time, calculate the values of the cost function for all the particle positions. 

If there is a change in $p_{j}$, then the attractor for the $j^{th}$ particle is recalculated using \eqref{eq5} while the amplitudes and phase values are recalculated by resetting the time for that particular particle as zero using \eqref{eq6}, \eqref{eq8} \& \eqref{eq9}. 
If instead there is a change in $g$, then all the attractors are changed accordingly using \eqref{eq5}. 
The amplitudes and phases are again recalculated using \eqref{eq6}, \eqref{eq8} \& \eqref{eq9} by resetting the time as zero for all particles.

This procedure, however, might in some cases lead to a significant loss of energy for the particle that found the new best position. This can be seen in an example when the best position is found at the boundary of the oscillation region where the velocity of the particle is close to zero. If that position is the new attractor, the potential energy of the string will be zero as well, as it freezes the particle until a new global best position is found. This is naturally not a desirable situation, as that particle is "punished" for being successful. To deal with this issue, we postulate a condition that the newly calculated amplitude is never smaller than the previous one -- more generally, finding a new best position can never lead to decrease of the energy in the system. 

Given enough time without finding a new global best, there is a possibility that the amplitude of particles whose personal best is far away from the global best becomes so small that the particle fails to effectively search the space between them, causing it to virtually become stuck in the middle, which might be a very unfavorable region. Numerical simulations  did show that this basically leads to an effective loss of this kind of particle, as it will almost never contribute with new results. A naive approach to solve this problem is by reducing the damping parameter $\lambda$, however the drawbacks of this is that this may also reduce the convergence in the desired cases. 

\begin{figure}[h]
    \centering
    \includegraphics[width=0.5\textwidth]{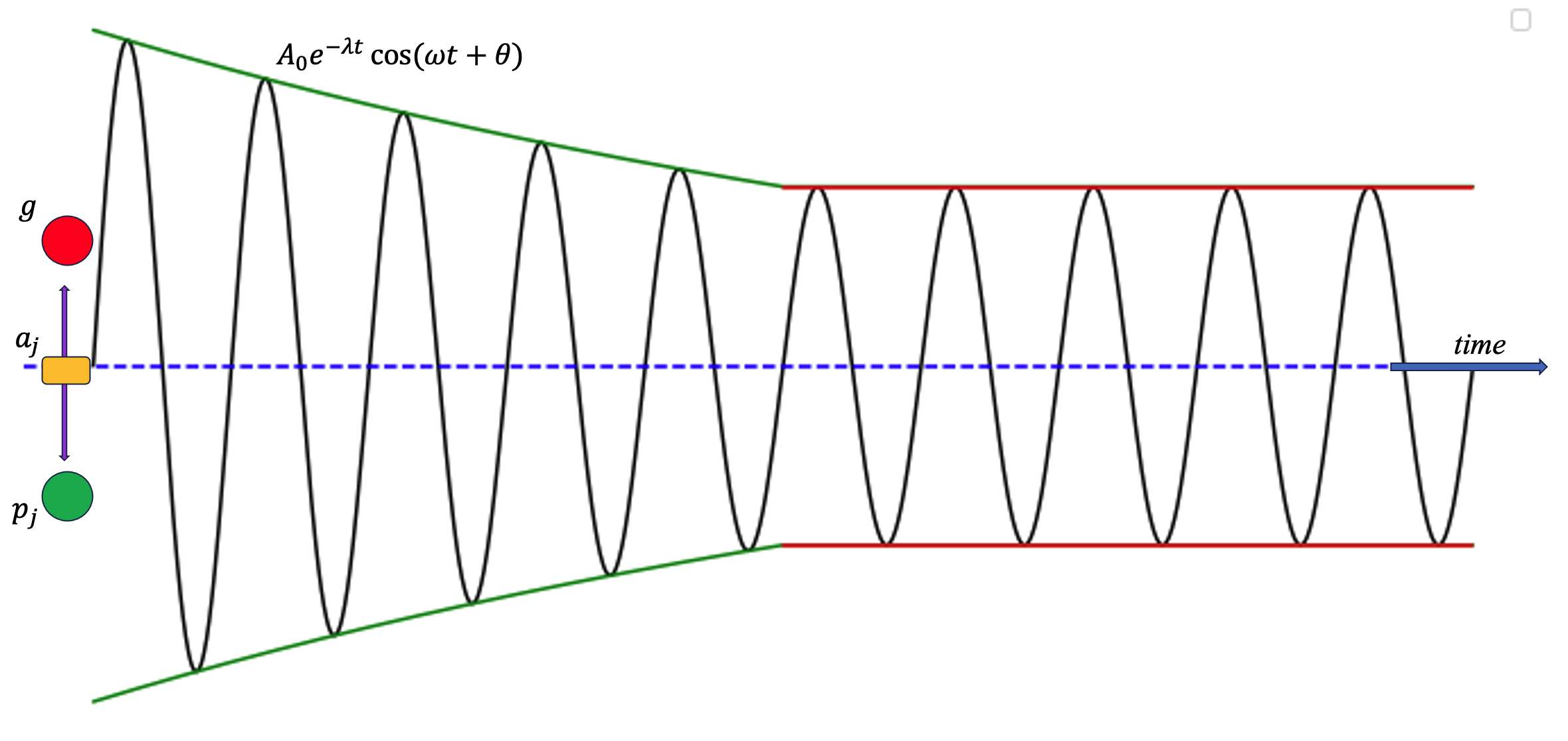}
    \caption{HOPSO Visualization: In one-dimension, the particle $j$ oscillates about the attractor $a_j$ which is set half-way between its personal best ($p_j$) and the swarm's global best ($g$) based on the weighted average equation \eqref{eq5}. The damping is switched off when the amplitude decreases, in the depicted case, it is approximately twice ($m = 2.05$) the distance between the attractor and one of the best positions.}
    \label{fig:hopso_visualization}
\end{figure}

We therefore included a different approach to resolve this issue, namely, to limit the amplitude from below to some threshold amplitude $A_{th}$ and prevent it from further damping. Naturally, this treshold amplitute should be in the order of the distance between the global and personal best positions to allow for searching in a region encompassing both of them. 
Thus we define 

\begin{equation}\label{eq7}
    A_{th} = \frac{|p_{j,d}-g_{d}|}{2}*m,
\end{equation}
where $m$ is a free parameter and define 

\begin{equation} \label{eq8}
    (A_{0})_{i+1} = \max((A)_{i},(A_{0})_{i+1}, A_{th}).
\end{equation}
A pictorial representation of \eqref{eq8} is shown in the Fig. \ref{fig:hopso_visualization}.

A  pseudo-code formalizing all the previously described procedures is shown in Algorithm \ref{pseudocode1}.

\begin{algorithm}[tb]
\caption{Harmonic Oscillator based Particle Swarm Optimization (HOPSO)}
\KwIn{Problem dimensions, Objective function, Algorithm parameters}
\KwOut{Candidate Optimal solution}
Set constants $c_1$, $c_2$, $\lambda$, $m$ for attraction weights and damping and minimal amplitude\;
Initialize particles with random positions $x_{j,d}$ and velocities $v_{j,d}$\;
Set initial personal best positions $p_i$ by starting positions for each particle\;
Choose initial global best position $g$\;
Calculate position of attractors: $a_j = \frac{c_1p_j+c_2g}{c_1+c_2}$\;
Calculate initial amplitude: $A_0 = \sqrt{(x(0) - a)^2 + \left(\frac{v(0)+\lambda(x(0)-a)}{\omega}\right)^2}$\;
Calculate initial phase: $\theta = \arccos \left(\frac{x(0)-a}{A_0}\right)$\;
\While{iteration $<$ max\_iterations}{
    \ForEach{particle}{
        \ForEach{dimension}{
            $A = \max(A_0e^{-\lambda t}, \frac{|p_j-g|}{2} \cdot m)$\;
            $x(t) = A_0e^{-\lambda t} \cos(\omega t + \theta) + x_{0}$\;
            $v(t) = -\omega(A_0e^{-\lambda t} \sin(\omega t + \theta)) - \lambda x(t)$\;
        }
    }
    \ForEach{particle}{
        Calculate Cost function from positions\;
        \If{$Cost\_function(x_{j,d}(t)) < Cost\_function(p_j)$}{
            Update $p_j$, best value, time, attractors, amplitude, phase\;
        }
    }
    \If{personal best value $<$ global best energy}{
        Update global best value and $g$\;
        \ForEach{particle and dimension}{
            Reset time, recalculate attractors, amplitude, phase\;
        }
    }
    iteration $\gets$ iteration $+$ 1\;
}
\label{pseudocode1}
\end{algorithm}

\section{Results}
\label{sec:guidelines}

We demonstrate now the performance of our algorithm on on a large set of commonly used test functions for multi-variable test functions.
These functions are listed in Tab. \ref{tab:Table1}.

These particular functions were chosen for their different and diverse properties, making them ideal for testing optimization algorithms \cite{OptimizationFunctions} and are a common, standard choice. The Ackley and Rastrigin functions have many local minima. Similarly, the Levy function's complex landscape challenges algorithms to avoid local minima. The Sphere function is a simple, unimodal, bowl-shaped function, while the Beale function is multimodal with sharp peaks at the corners of the input domain. The Goldstein-Price function is highly multimodal and complex, while the Schwefel function, known for its large search space, presents numerous traps. The Rosenbrock function's narrow, curved valley tests precision, while the Drop-Wave function features steep drops and peaks. The Cross-in-Tray function and the Michalewicz function, with their deep valleys and sharp peaks, also make it extremely challenging for optimization algorithms to find the optimum and are therefore suitable choices as well. Each function was chosen to uniquely test different aspects of the optimization algorithm that we present in this text. 

%

\begin{table*}[htbp]
    \caption{Commonly used test functions for optimization methods}
    \centering
    \renewcommand{\arraystretch}{3} 
    
    \begin{tabular}{c p{6.5cm} c c c c}
    \textbf{Name} & \textbf{Functional Form} & \textbf{Modality} & \textbf{Initial range} & \textbf{F\textsubscript{min}} & \textbf{Dimension} \\
    \hline
    Ackley & $\begin{aligned}-a \exp\left(-b \sqrt{\frac{1}{d} \sum_{i=1}^{d} x_i^2}\right) \\ - \exp\left(\frac{1}{d} \sum_{i=1}^{d} \cos(c x_i)\right) + a + \exp(1)\end{aligned}$ & Multimodal & {[}-32.76,32.76{]} & 0 & 10\\
    \hline
    Beale & $\begin{aligned}(1.5 - x_1 + x_1 x_2)^2 + (2.25 - x_1 + x_1 x_2^2) \\ + (2.625 - x_1 + x_1 x_2^3)\end{aligned}$ & Multimodal & {[}-5,5{]} & 0 & 2 \\
    \hline
    Cross-in-Tray & $\begin{aligned} -0.0001 [ & |\sin(x_1) \sin(x_2) \\ & \exp(|100 - \frac{\sqrt{x_1^2 + x_2^2}}{\pi}|)| + 1]^{0.1} \end{aligned}$ & Multimodal & [-10,10] & -2.06261 & 2 \\
    \hline
    Drop-Wave & $-\frac{1 + \cos(12 \sqrt{x_1^2 + x_2^2})}{0.5 (x_1^2 + x_2^2) + 2}$ & Multimodal & {[}-5.12,5.12{]} & -1 & 2\\
    \hline
    Goldstein-Price & $\begin{aligned}[1 + (x_1 + x_2 + 1)^2 (19 - 14 x_1 + 3 x_1^2 - 14 x_2 \\ + 6 x_1 x_2 + 3 x_2^2)] \cdot [30 + (2 x_1 - 3 x_2)^2 (18 \\ - 32 x_1 + 12 x_1^2 + 48 x_2 - 36 x_1 x_2 + 27 x_2^2)]\end{aligned}$ & Multimodal & {[}-2,2{]} & 3 & 2 \\
    \hline
    Griewank & $\frac{1}{4000} \sum_{i=1}^{d} x_i^2 - \prod_{i=1}^{d} \cos\left(\frac{x_i}{\sqrt{i}}\right) + 1$ & Multimodal & {[}-600,600{]} & 0  & 10 \\
    Levy & $\begin{aligned}\sin^2(\pi w_1) + \sum_{i=1}^{d-1} (w_i - 1)^2 [1 + 10 \sin^2(\pi w_i + 1)] \\ + (w_d - 1)^2 [1 + \sin^2(2 \pi w_d)]\end{aligned}$ & Multimodal & {[}-10,10{]} & 0 & 10\\
    \hline
    Michalewicz & $-\sum_{i=1}^{d} \sin(x_i) \left[\sin\left(\frac{i x_i^2}{\pi}\right)\right]^{2m}$ & Multimodal & {[}0,$\pi${]} & -4.687 & 5\\
    \hline
    Rastrigin & $10d + \sum_{i=1}^{d} [x_{i}^2 - 10 \cos(2 \pi x_{i})]$ & Multimodal & {[}-5.12,5.12{]} & 0  & 10\\
    \hline
    Rosenbrock & $\sum_{i=1}^{d-1} [100 (x_{i+1} - x_i^2)^2 + (x_i - 1)^2]$ & Unimodal & {[}-5,10{]} & 0 & 10\\
    \hline
    Schwefel & $\sum_{i=1}^{d} [-x_i \sin(\sqrt{|x_i|})]$ & Multimodal & {[}-500,500{]} & 0 & 10\\
    \hline
    Sphere & $\sum_{i=1}^{d} x_{i}^2$ & Unimodal & {[}-10,10{]} & 0 & 5 \\
    \end{tabular}
    \label{tab:Table1}
\end{table*}

The performance of HOPSO will be compared to that of the standard PSO, COBYLA \cite{Powell1994}, and Differential Evolution (DE) \cite{storn1997differential} optimization methods.

Constrained optimization by linear approximation (COBYLA) is a gradient-free simplex based optimization method.
It was first introduced by Michael J.D. Powell in 1994 \cite{Powell1994}. To describe the algorithm briefly, it operates by creating a simplex, a polytope of \(n+1\) vertices for an \(n\)-dimensional space, and using the values of the objective function at the vertices of this simplex it approximates the objective function along with linear contraints, solving linear programming problems within a trust region.
The simplex and the trust region are adjusted iteratively until the convergence is obtained. For more details refer to Powell’s original paper \cite{Powell1994}.

Differential Evolution (DE) is an another meta-heuristic algorithm like PSO.
DE is an optimization technique that begins with a set of possible solutions and gradually refines them.
DE was first given by Rainer Storn and Kenneth Price in 1997 \cite{storn1997differential}.
To briefly describe DE, a population of candidate solutions is initialized randomly. 
For each candidate, a mutant vector is generated by adding the weighted difference between two randomly selected population vectors to a third vector. 
This mutant vector is then recombined with the target vector, and a selection process determines whether the new vector replaces the target vector based on a fitness evaluation. 
This process is repeated iteratively until a stopping criterion is met.  It was shown recently \cite{piotrowski2023particle} that DE, although not being  as popular as PSO, outperforms PSO in many cases, hence it is a natural choice for an optimizer for comparison.

To compare optimization algorithms, it is essential to define a "budget" based on a specific resource. Using the maximum number of iterations as the budget can be problematic since different optimizers may use varying numbers of function calls per iteration, even if they share the same maximum iterations. Therefore, in our case, we select function evaluations as the budget. This budget represents how many times the algorithm can query the objective function to assess its performance. By using function evaluations, we ensure a fair comparison across algorithms regardless of their internal operations.

In principle, for a given  budget, the same method can achieve different results for different values of tuning parameters, so we use the  commonly-used ``standard settings'' which result in a typically-good performance for each optimization method. Specifically, here in this study Scipy's optimization module \cite{2020SciPy-NMeth} was used without altering its standard settings for each optimizer to implement COBYLA and DE. PSO was run via our own implementation but using the standard parameter settings found in the literature mentioned earlier.

 Depending on the complexity of the problem, we chose our budget to be either 1000 or 10000 function evaluations.
 The PSO's chosen parameter-values are those that are widely accepted based on empirical studies $\chi=0.729$ and $c_1$, $c_2$ being each 2.05 \cite{4223164}. 
 The HOPSO settings are designed to mirror the equivalent parameters in PSO, ensuring a fair playing field between the optimizers. The HOPSO settings are $c_1=c_2=\omega =1 $, $t_{ul} = 2\pi$ and $m = 2.05$. 

The crucial difference which offers the superior advantage of HOPSO against PSO is the tunability of the damping parameter $\lambda$. Here we provide a general guideline (or starting point) for selecting $\lambda$ that may be adjusted by the user as needed. As  $\lambda$ governs damping, it naturaly should be inversely proportional to the budget per particle. For a higher budget of function evaluations per particle, there is less need for damping in the system -- since less damping allows for a broader search, and vice versa. As a proportionality constant, we define a scaling factor $s$. The relationship for setting the $\lambda$ based on the budget is as follows
\begin{equation} \label{eq9}
\lambda \equiv s \left( \frac{B}{N} \right)^{-1}
\end{equation}
where $B$ is the number of function evaluations (i.e. budget), $N$ is the total number of particles, and $s$ is a scaling factor. Extremely large or small values of $s$ are undesirable, as they lead to overly strong or weak damping, respectively. 

For the case where $s=1$, if no updates occur during the optimization, the amplitude $A$ reduces to $A/e$ (approximately $0.36A$) over the total time period of $B/N$. With updates and due to the reduced damping in certain cases, the final amplitude might be even higher, resulting in a broad, but not very precise search. 

A more aggressive damping can be achieved by choosing $s=10$, in which case, within the same time period, the amplitude decreases to $0,000045A$) (or less due to updates) allows for a more precise search at a higher risk to fall into a local minimum. In the first part of our study we did choose $10$ for all the functions to keep a fair battlefield. In the second part, we elaborate on the influence of $\lambda$ in specific cases. 

\begin{figure*}
	\begin{subfigure}{0.48\linewidth}
		\includegraphics[width=\linewidth]{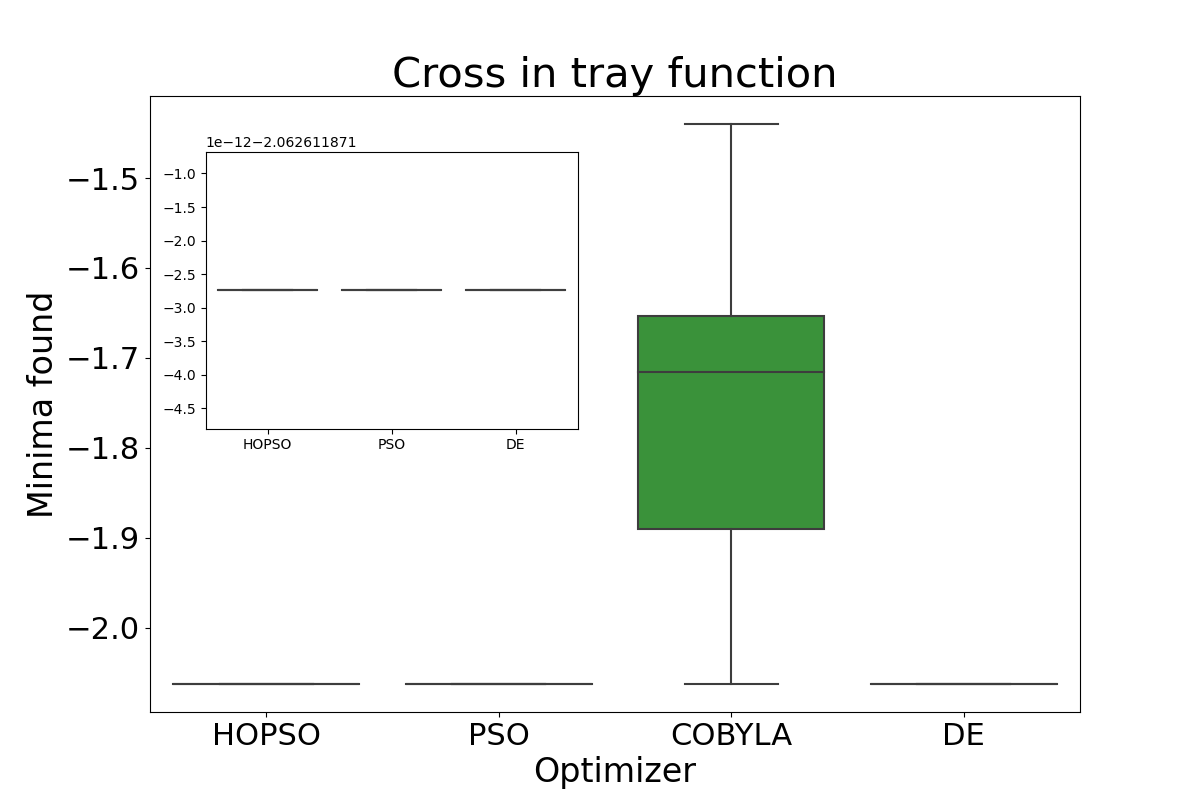}
		\caption{Except for COBYLA, all other optimizers converge to the minima, with HOPSO showing the highest precision (not apparent in the figure).}
		\label{}
	\end{subfigure}
 \hfill
	\begin{subfigure}{0.48\linewidth}
		\includegraphics[width=\linewidth]{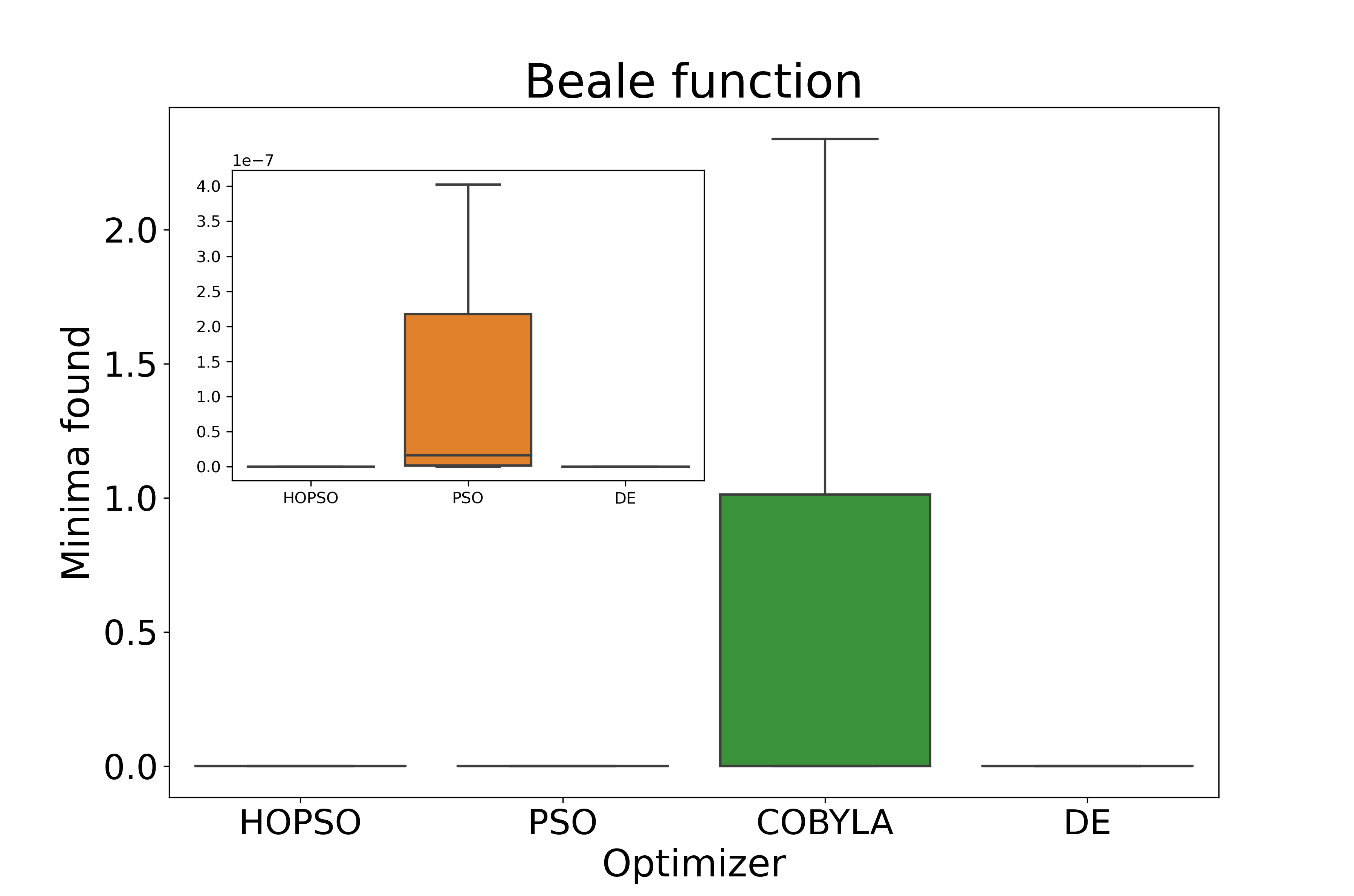}
		\caption{All optimizers except for COBYLA converge reasonably well, with PSO and COBYLA failing to reach the precision of HOPSO.}
		\label{}
	\end{subfigure}
    \vfill
    \begin{subfigure}{0.48\linewidth}
        \includegraphics[width=\linewidth]{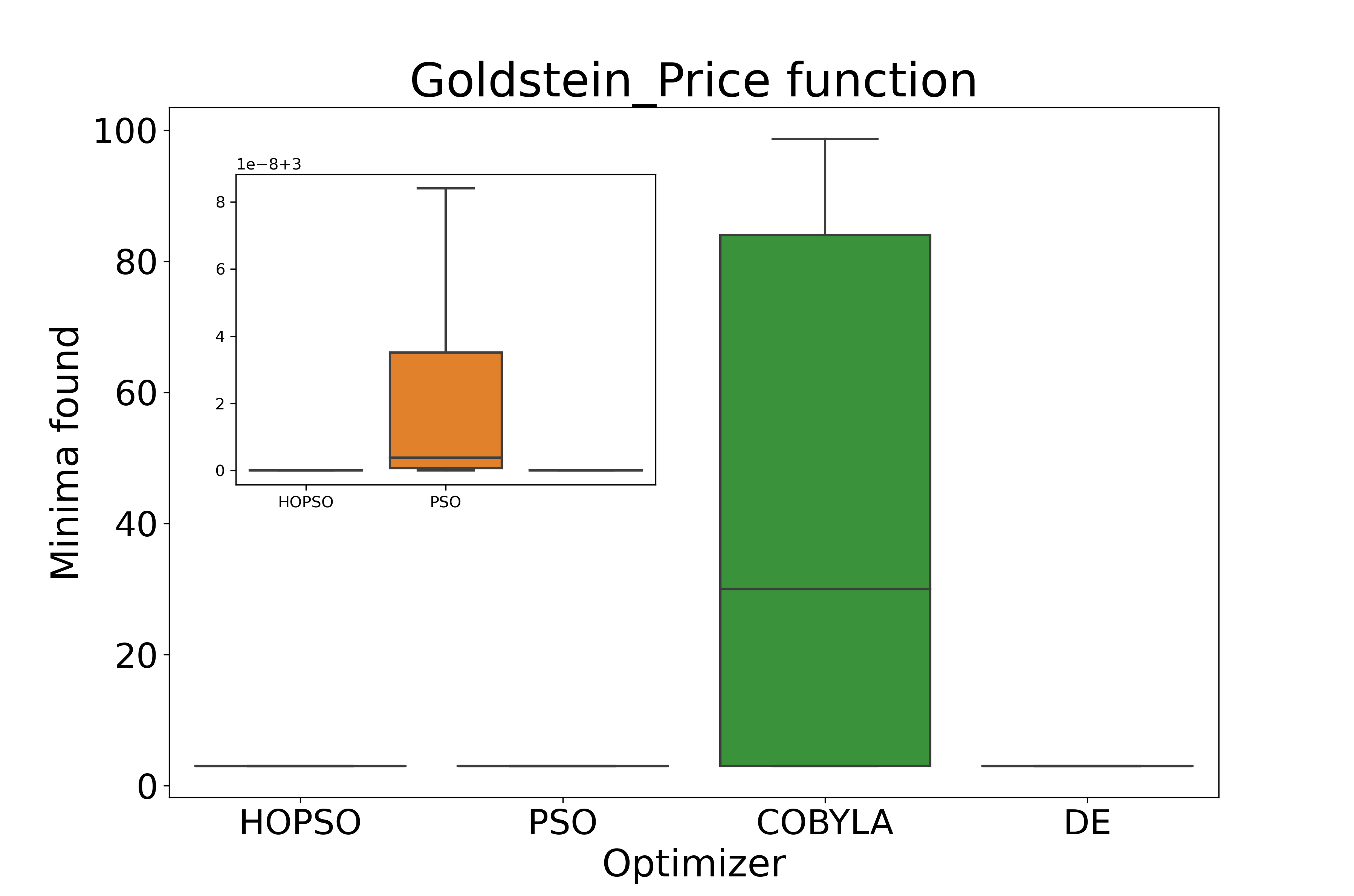}
        \caption{Here COBYLA and PSO do not converge to the optimum unlike DE and HOPSO, with the latter outperforming its rival counterparts in precision.}
        \label{}
    \end{subfigure}
    \hfill
    \begin{subfigure}{0.48\linewidth}
        \includegraphics[width=\linewidth]{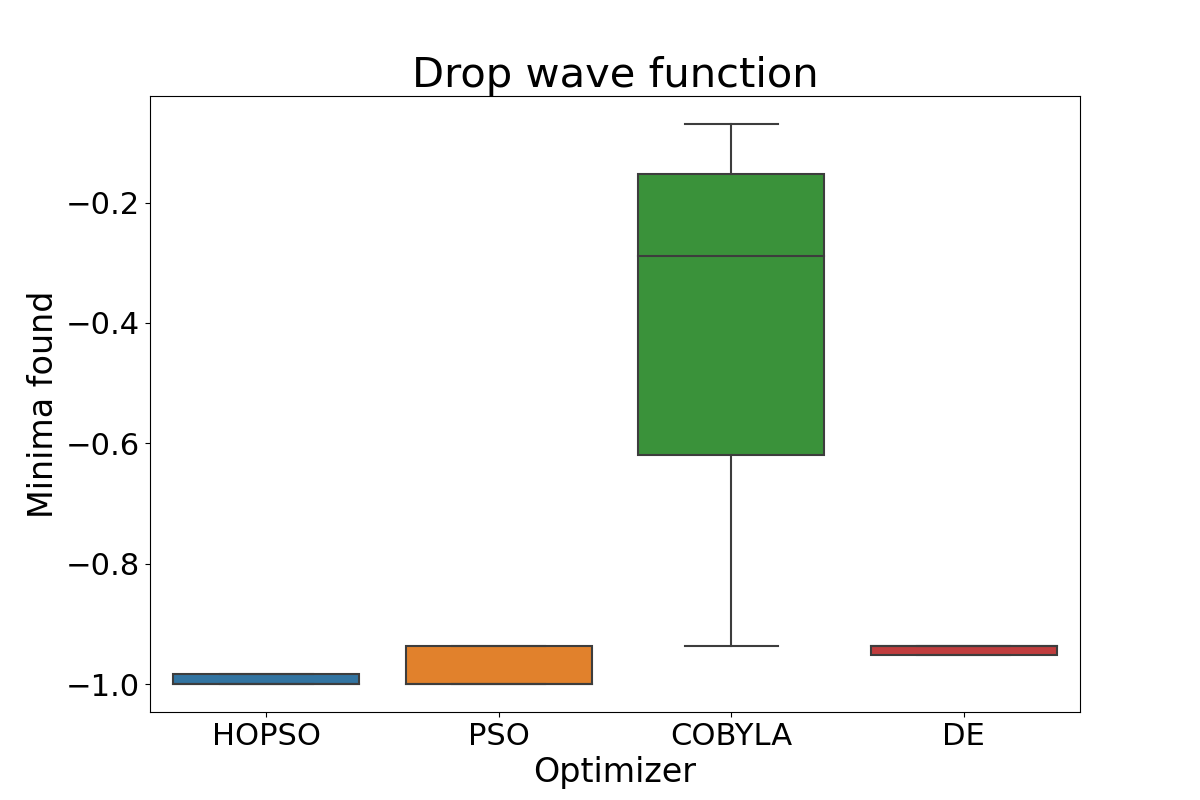}
        \caption{Unlike in previous cases, none of the optimizers converges perfectly within a given budget. Here, both PSO and HOPSO do reach better results than COBYLA and DE, with HOPSO achieving higher precision.}
        \label{}
    \end{subfigure}

        \caption{\raggedright Comparison of the performance of different optimizers on constant dimensional ($dimension = 2$) test functions.
        }
        \label{fig:constant_dimension_results}
\end{figure*}
\begin{figure*}
	\centering
	\begin{subfigure}{0.48\linewidth}
		\includegraphics[width=\linewidth]{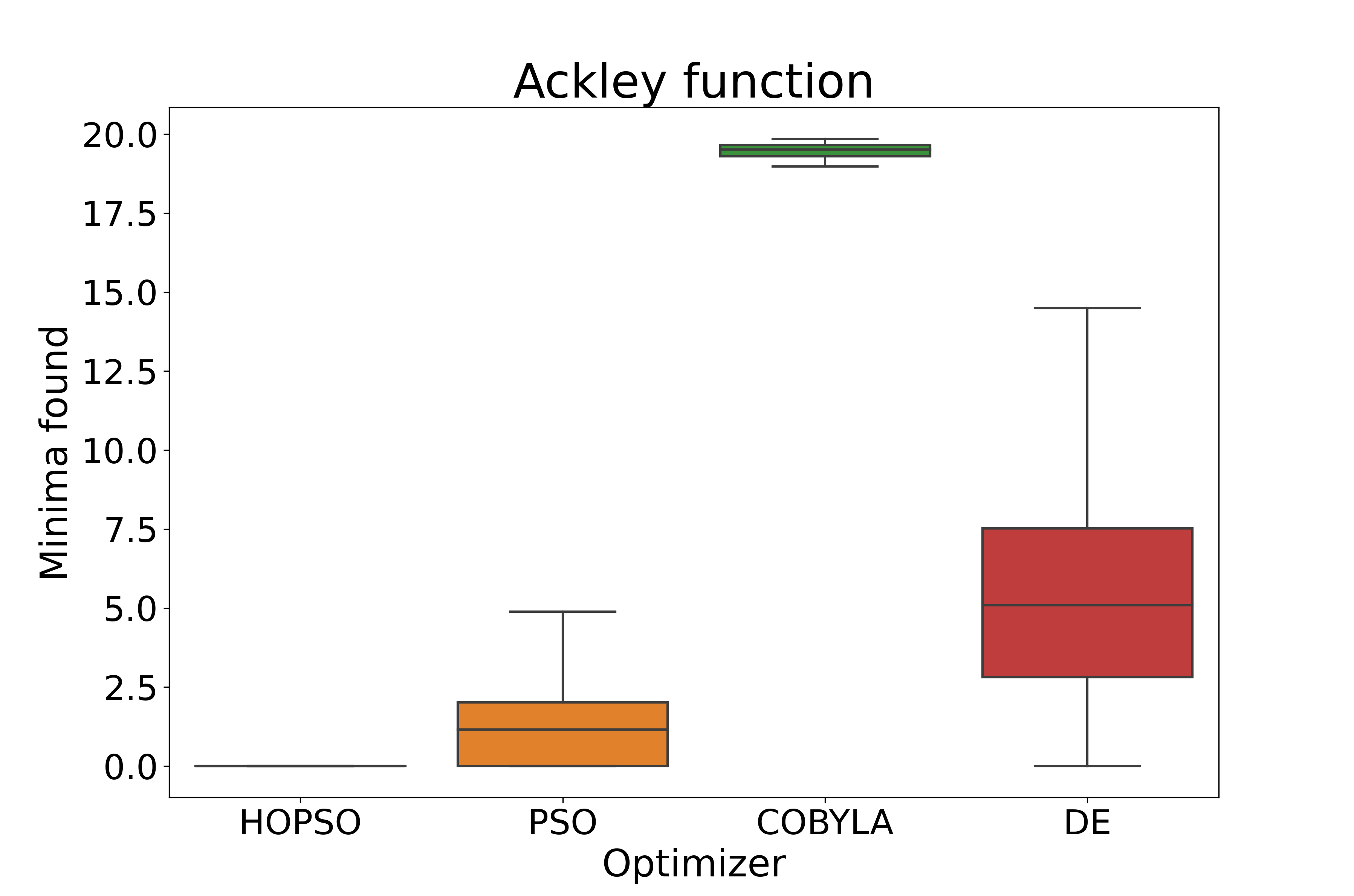}
		\caption{All optimizers except HOPSO fail to converge to the minima, with HOPSO being more precise and more accurate than its competitors. }
		\label{}
	\end{subfigure}
 \hfill
	\begin{subfigure}{0.48\linewidth}
		\includegraphics[width=\linewidth]{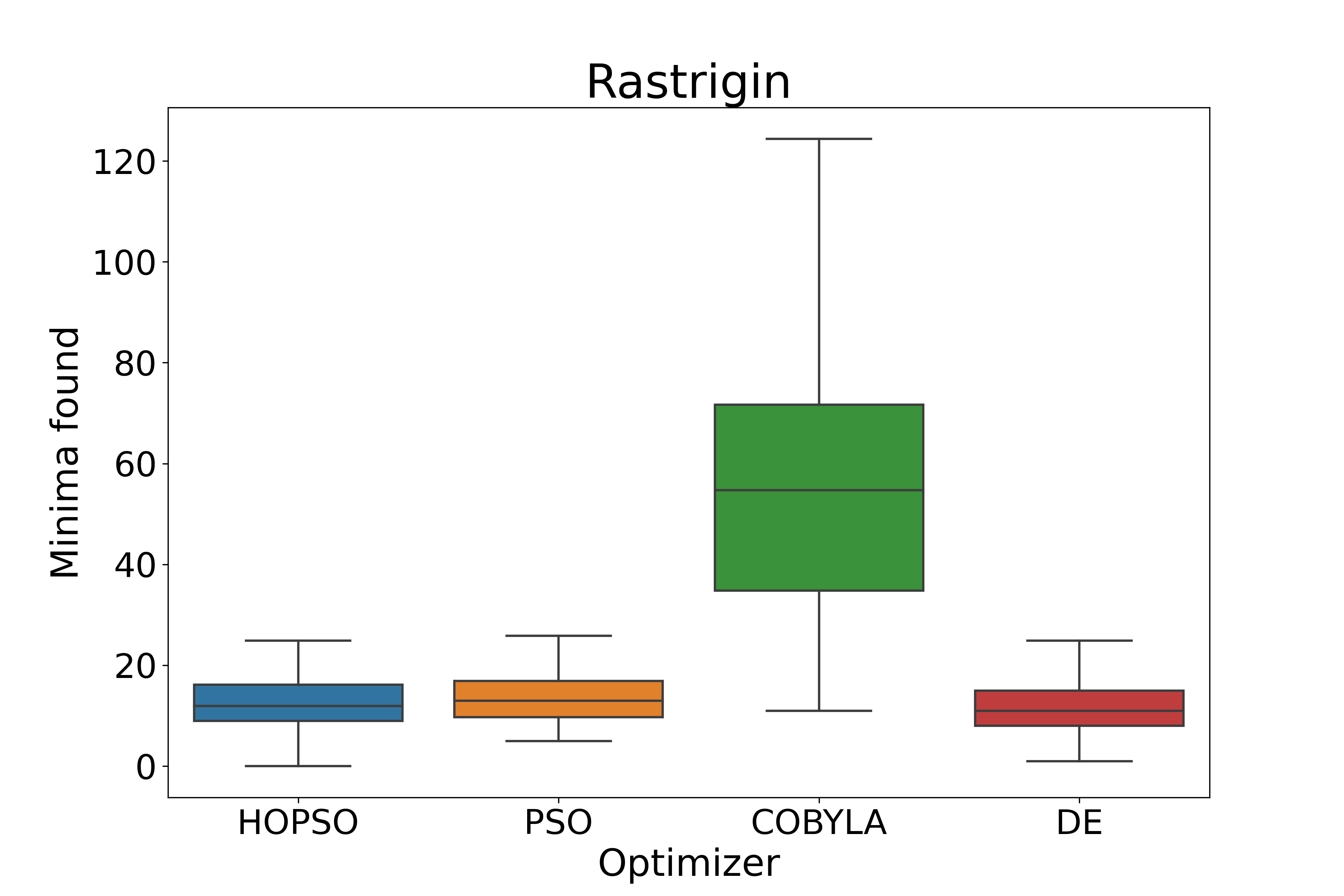}
		\caption{All optimizers fail to converge to the minima. However, DE performs the best slightly ahead HOPSO and PSO which in turn are significantly better than COBYLA.}
		\label{}
	\end{subfigure}
    \vfill
    \begin{subfigure}{0.48\linewidth}
        \includegraphics[width=\linewidth]{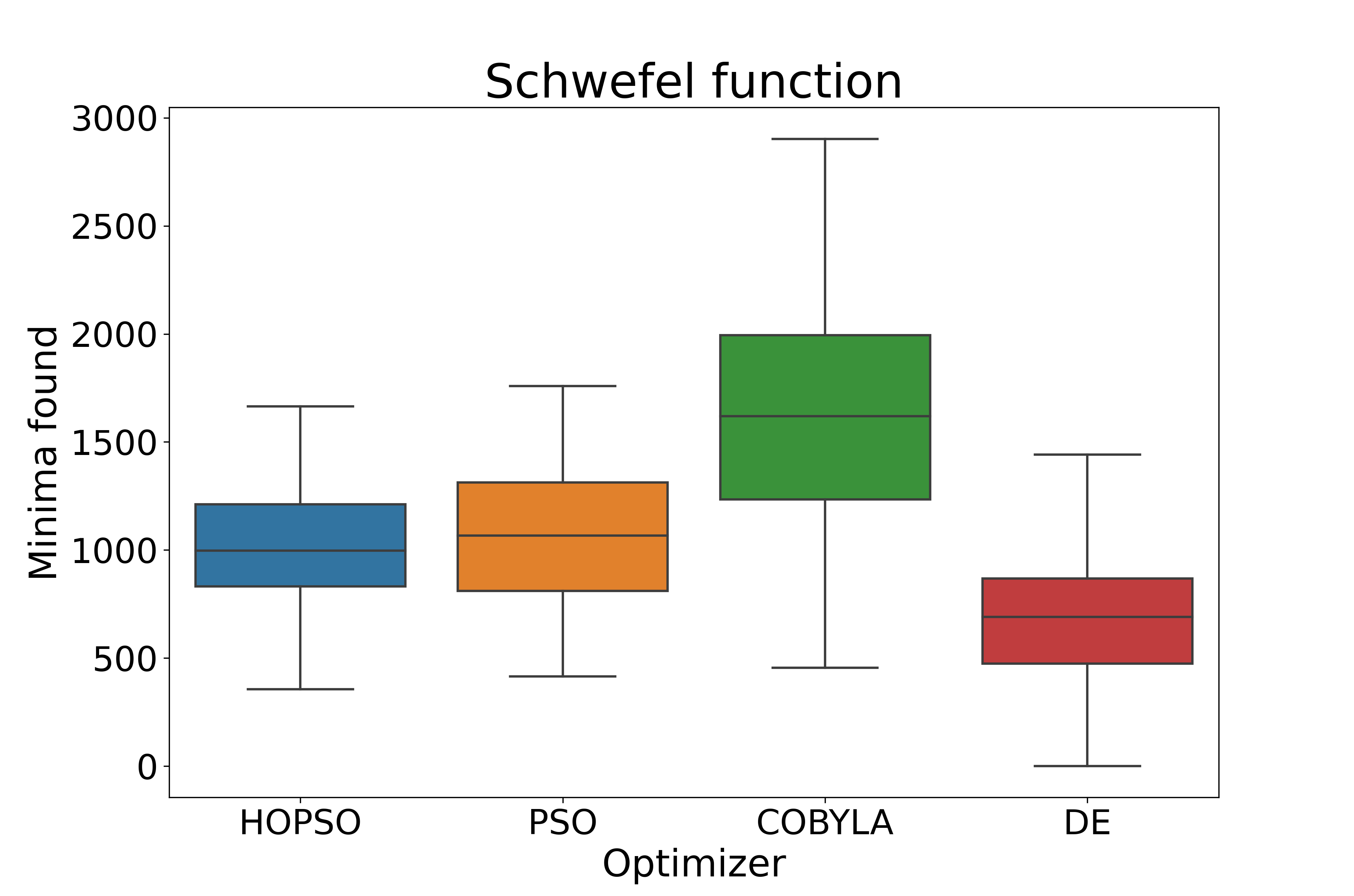}
        \caption{All optimizers fail to converge to the minima by a significant amount. Among these, DE performs the best, followed by HOPSO and PSo. }
        \label{}
    \end{subfigure}
    \hfill
    \begin{subfigure}{0.48\linewidth}
        \includegraphics[width=\linewidth]{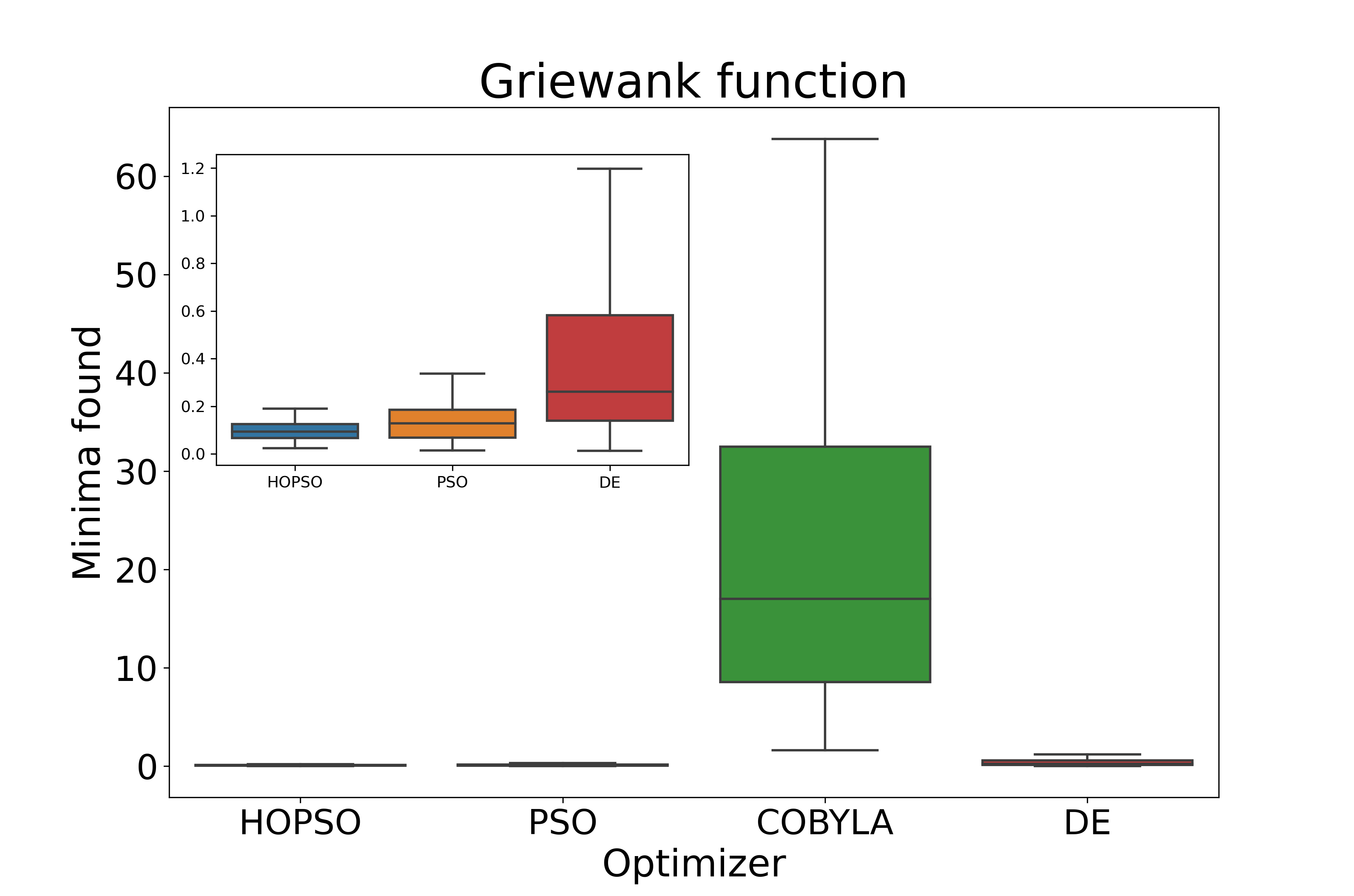}
        \caption{ HOPSO and PSO perform better than the other two competitors by being more precise and accurate. As can be seen from the figure, HOPSO is more precise and has a lower median than PSO. }
        \label{}
    \end{subfigure}
    \vfill
    \begin{subfigure}{0.48\linewidth}
        \centering
        \includegraphics[width=\linewidth]{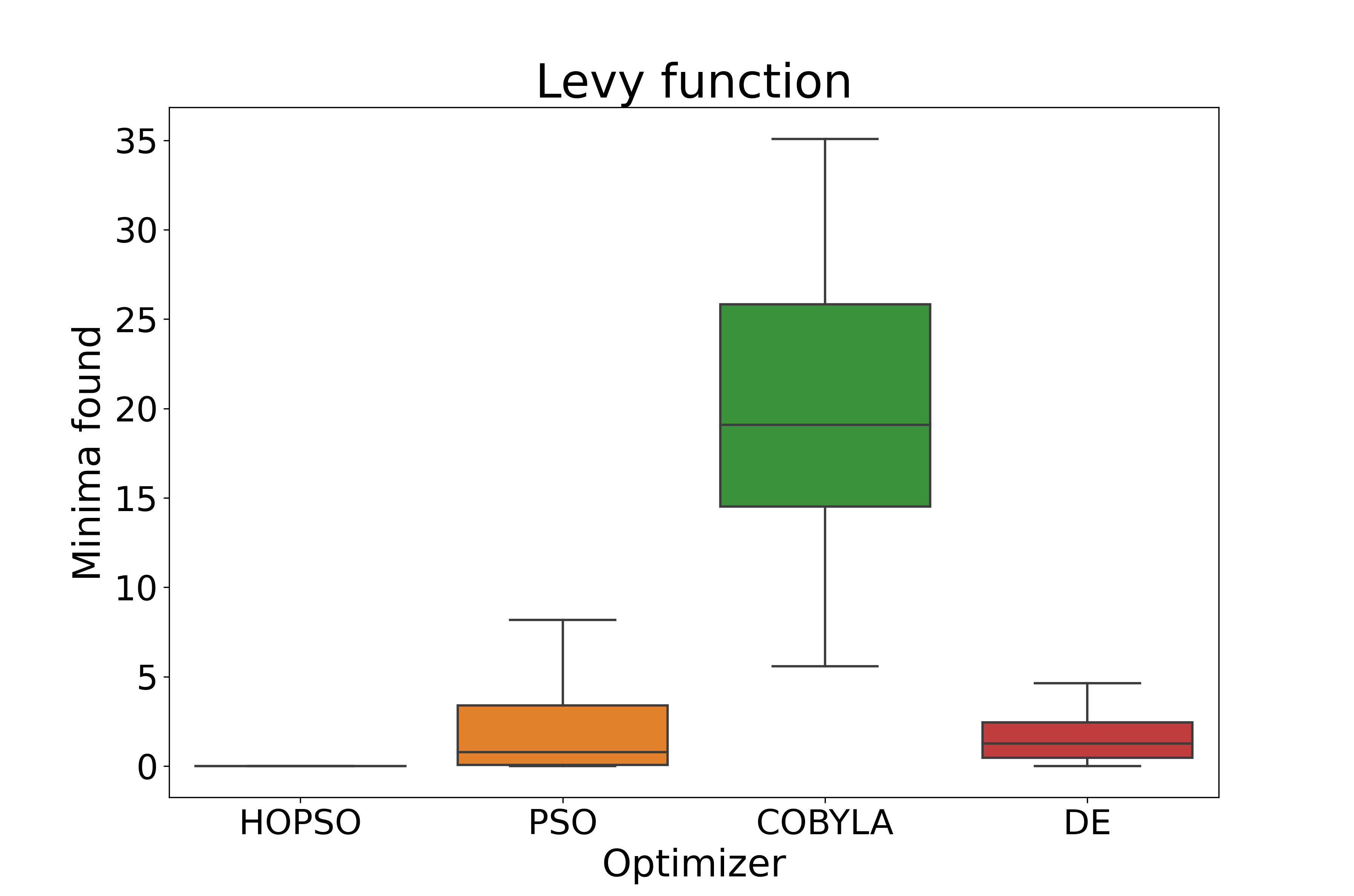}
        \caption{Except for HOPSO, all optimizers fail to converge to the minima. HOPSO performs the best followed by PSO, DE and COBYLA.}
        \label{}
    \end{subfigure}
    \hfill
    \begin{subfigure}{0.48\linewidth}
        \centering
        \includegraphics[width=\linewidth]{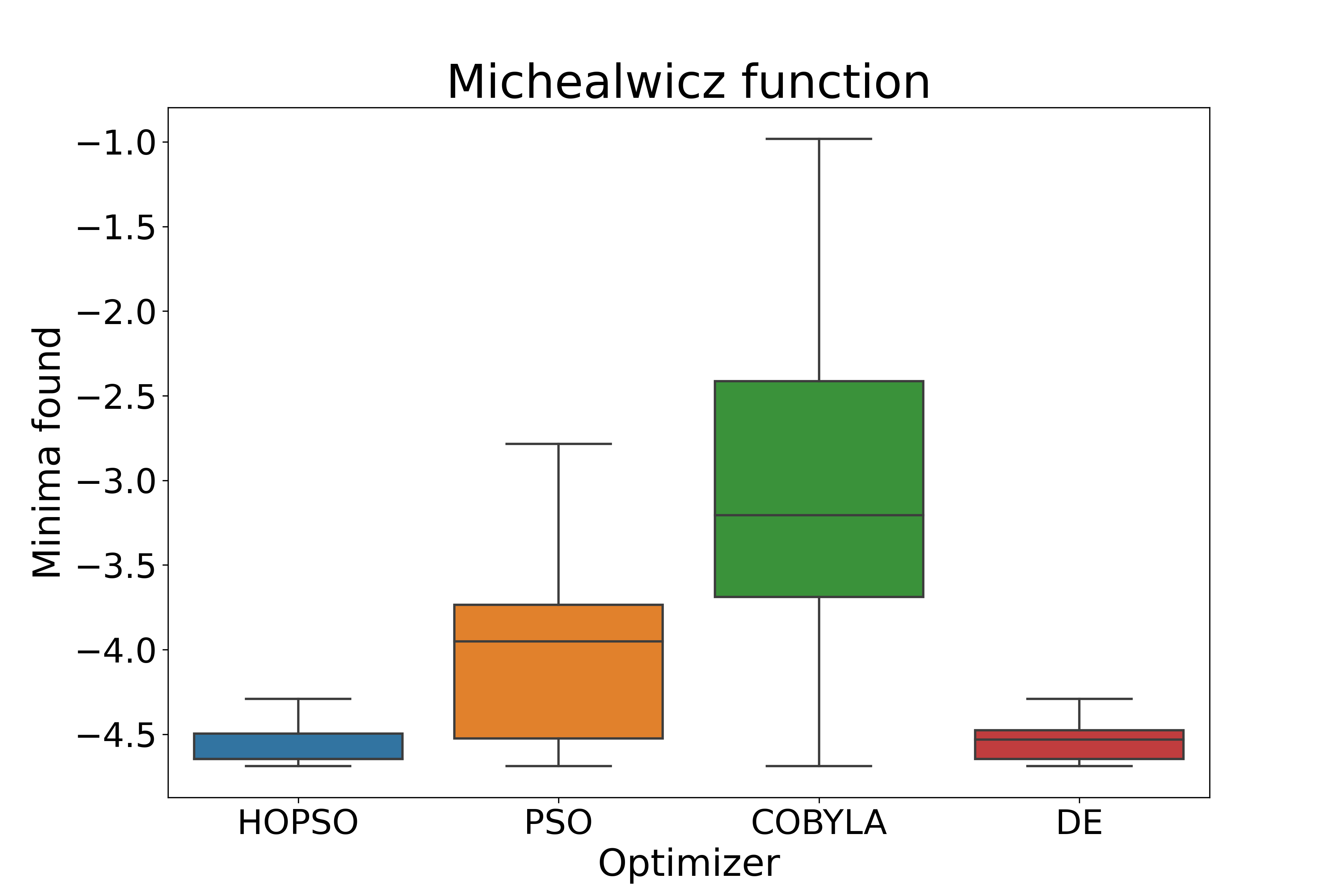}
        \caption{ HOPSO and DE show the most consistent performance with minimal spread. DE performs the best, slightly ahead of HOPSO, which significantly outperforms both PSO and COBYLA.}
        \label{}
    \end{subfigure}

        \caption{\raggedright A comparison of the performance of different optimizers on varying-dimensional test functions. All test functions here are chosen to have a $dimension = 10$ except for Michealwicz, whose $dimension$ is chosen to be $5$. 
        }
        \label{fig:inconstant_dimension_results}
\end{figure*}
\begin{figure*}
	\centering
	\begin{subfigure}{0.49\linewidth}
		\includegraphics[width=\linewidth]{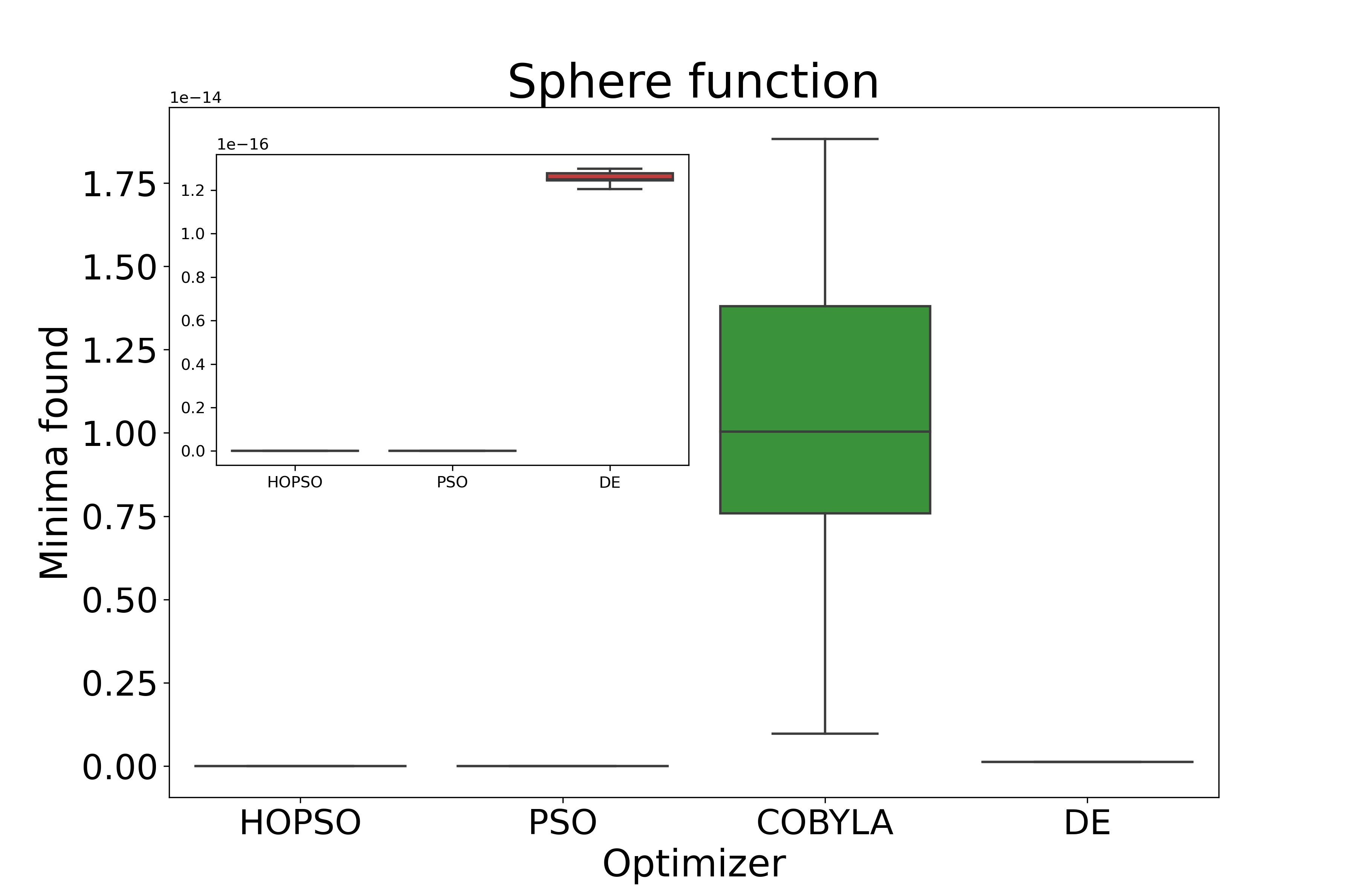}
		\caption{All optimizers converge to the minima. For higher precision scenarios, HOPSO and PSO perform the best.}
		\label{Sphere}
	\end{subfigure}
 \hfill
	\begin{subfigure}{0.49\linewidth}
		\includegraphics[width=\linewidth]{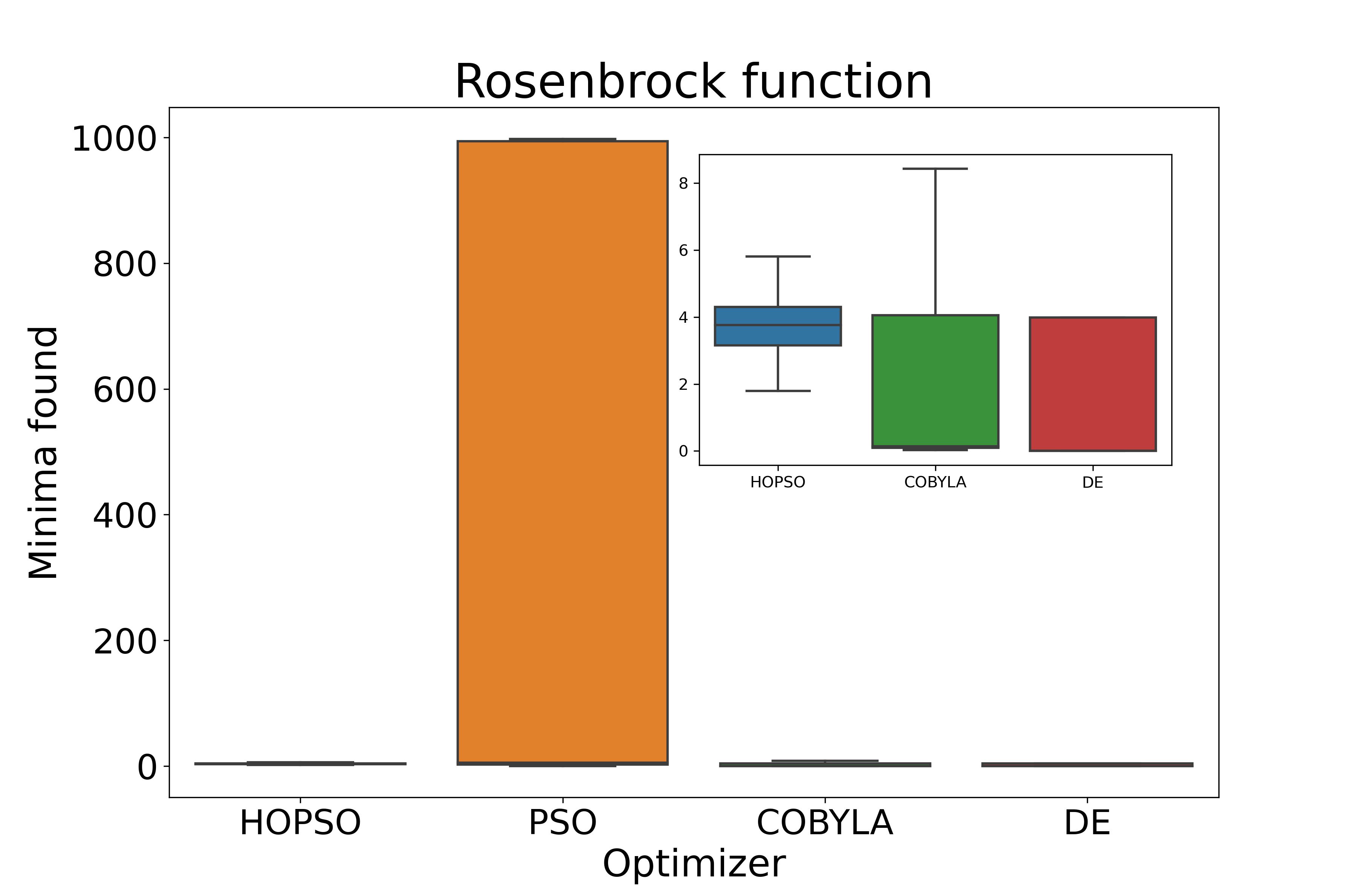}
		\caption{None of the optimizers successfully converge to the minima. However, DE and COBYLA outperform both PSO and HOPSO, with PSO showing the weakest performance. }
		\label{Rosenbrock}
	\end{subfigure}
    
	\caption{\raggedright A comparison of the performance of four different optimizers on Unimodal test functions.   }
	\label{fig:unimodal_results}
\end{figure*}

The results presented in the Figs. \ref{fig:constant_dimension_results},\ref{fig:inconstant_dimension_results} \& \ref{fig:unimodal_results} compare the performance of HOPSO, PSO, COBYLA, and DE on a diverse set of test functions. 
It is important to note that all the box plots shown do not consider the outliers. As seen in Fig. \ref{fig:constant_dimension_results}, for the two-dimensional test functions HOPSO outperforms the other optimizers in all cases;
COBYLA consistently underperforms, while DE ranks second.
For the Cross-in-Tray function, all optimizers except COBYLA reach the minimum.
The performance trends are similar for the Beale and Goldstein-Price functions.
For the Drop Wave function, HOPSO gets closest to the minimum, followed by PSO, DE, and COBYLA.

In Fig. \ref{fig:inconstant_dimension_results},  which shows test functions with varying dimensions, HOPSO excels on the Ackley, Levy, and Griewank functions.
Its performance is comparable to DE on the Rastrigin and Michalewicz functions. For the Schwefel function, none of the optimizers reach the minimum (with DE performing best, albeit still poor).
In Fig. \ref{fig:unimodal_results}, for the unimodal test functions, HOPSO, along with PSO and DE, converges to the minimum for the Sphere function. For the Rosenbrock function, HOPSO works significantly better then the original PSO, while it is slightly outperformed by DE and COBYLA.

All in all, HOPSO clearly outperforms the other three optimizers in six cases. In two cases, HOPSO, PSO, and DE perform equally well, while DE performs better in four cases, however in three of them the reslts of all optimizers are far away from the actual solution, so one might conclude that all of them did fail.

\begin{figure*}
	\centering
	\begin{subfigure}{0.48\linewidth}
		\includegraphics[width=\linewidth]{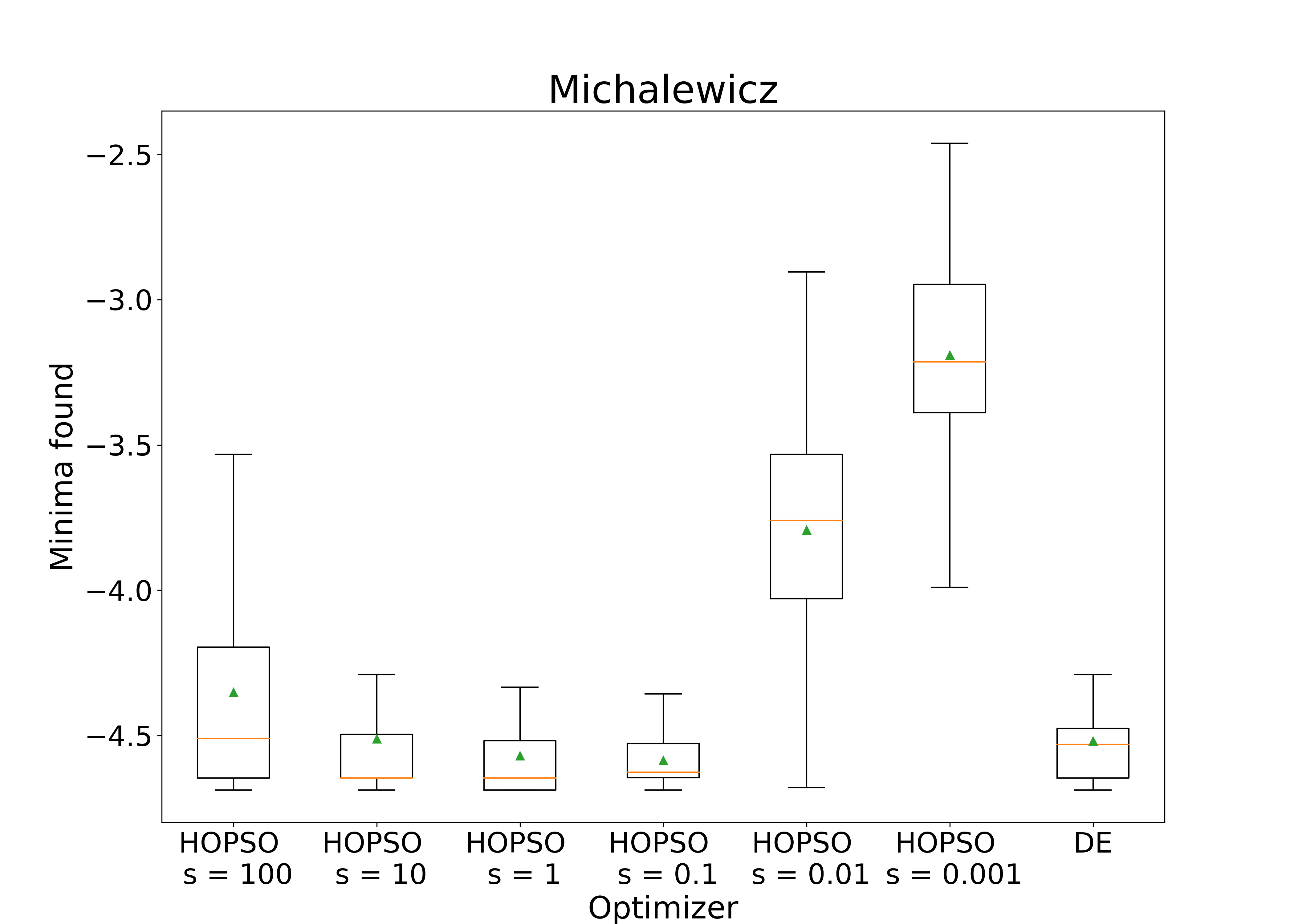}
		\caption{The results of HOPSO with scaling factor set to $0.1$ and $1$ outperform the previously best-performing optimizer, DE. }
		\label{}
	\end{subfigure}
 \hfill
	\begin{subfigure}{0.48\linewidth}
		\includegraphics[width=\linewidth]{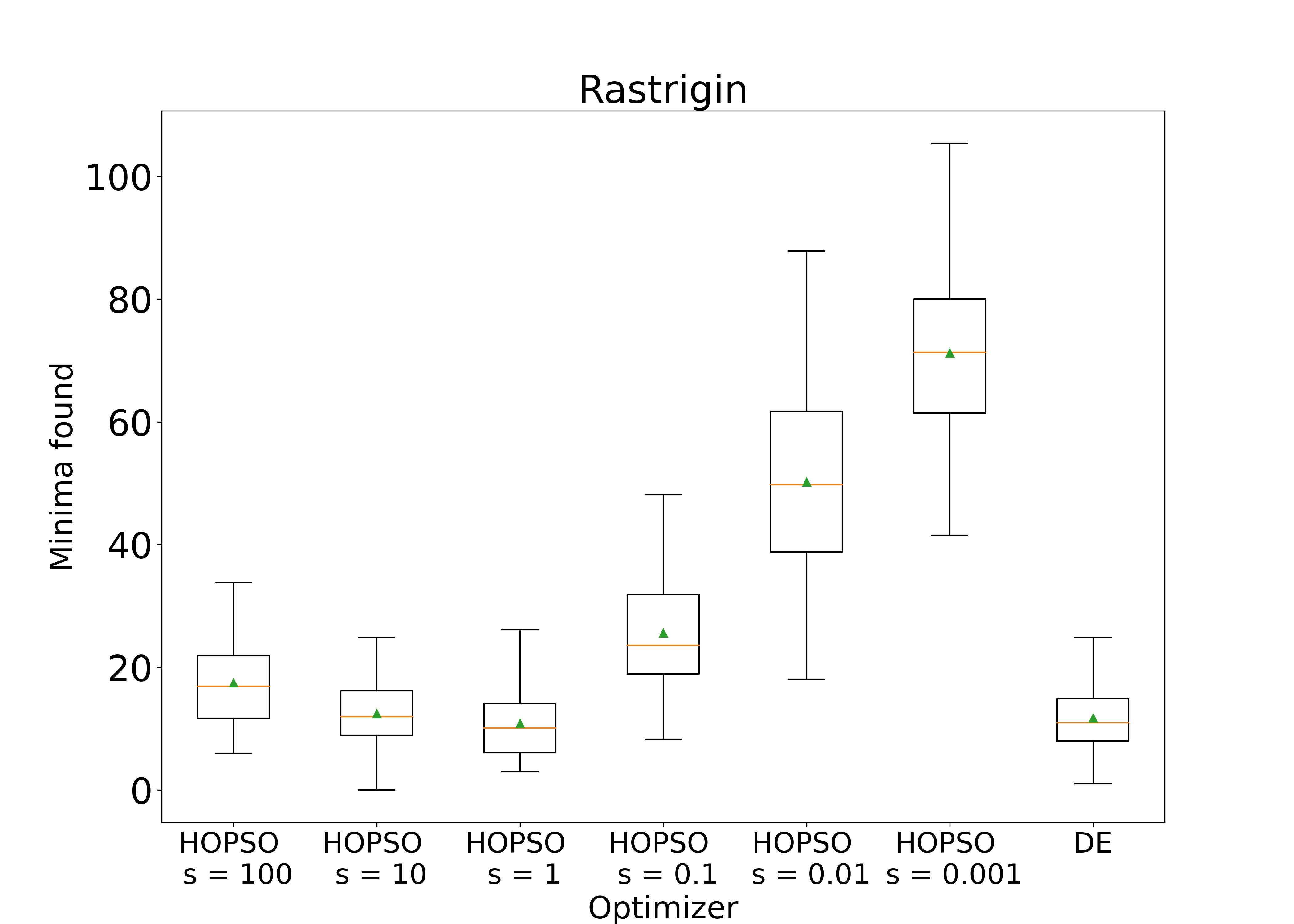}
		\caption{The results of HOPSO with the setting of the scaling factor $s$ to $1$ outperforms the previously best-performing optimizer, DE.}
		\label{}
	\end{subfigure}

	\caption{\raggedright Comparison of the performance of HOPSO with varying $\lambda$ based on the scaling factor $s$ with DE on Michalewicz and Rastrigin functions. The green triangle represents the mean of the distribution and the yellow line representing the median. It can be seen that by fine-tuning the scaling factor HOPSO outperforms DE. 
    }
	\label{fig:varying lambda}
\end{figure*}

These results for HOPSO were achieved without fine-tuning the hyperparameters, similar to the other optimizers. To research the abilities of HOPSO we did test it for different values of the $s$ parameter in cases where, in its original settings, it was outperformed by DE. For Michalewicz function the mean result got better from $-4.5119$ ($s=10$) to $-4.5860$ ($s=0.1$) and 
for the Rastrigin function from $12.458$ ($s=10$) to $10.841$ ($s=1$), in both cases outperforming the DE. Detailed results with varying scaling factor $s$ are shown in Fig. \ref{fig:varying lambda}.

\section{Conclusion}

In this paper, we suggested a new optimization method based on the well researched Particle swarm optimization (PSO) method by introducing the concept of conservation of energy to prohibit wild oscillations and velocity explosions. In the newly developed method we also introduced an independent tuning parameter that allows to adjust the convergence of the method depending on the expected number of function evaluations. 

We have shown the power of the new procedure through its application onto a set of standard benchmark test optimization functions and compared to the original PSO method, as well as other non-gradient methods including COBYLA and DE. 
In literally all the cases, HOPSO showed to be at least as good as the original PSO. Thus, in any case when one may consider using PSO for their optimization purposes, HOPSO merits consideration over the original PSO method. HOPSO also did outperform COBYLA in all cases.

In most cases, it has also outperformed the DE method. In some of the cases, it turned out that the tunability of HOPSO allows to get better results than DE. The few cases where DE proved to be better are associated with situations were none of the optimizers managed to get a reasonably good result close to the actual minimum -- in these cases the DE was able to scan a larger space than any standard method. 

In our future research, we foresee the application of HOPSO in high-dimensional optimization problems that involve significant number of parameters, in particular in connection with quantum computers. We believe that the simple tunability of the convergence of the method will simplify the search in these highly degenerate problems.


\begin{table*}
\centering
\renewcommand{\arraystretch}{1.5} 
\begin{tabular}{| p{3cm} | p{2cm}| p{2cm}| p{2cm}| p{2cm}| p{2cm}| p{2cm}|}
\hline
Function & Function evaluations & F\textsubscript{min} & \multicolumn{4}{c|}{Mean}  \\ \cline{4-7}
 & & & HOPSO & PSO & COBYLA & DE \\ 
 \hline
 Ackely & 10000 & 0 & \textbf{0.0115} & 1.3824 & 19.410 & 5.6180 \\
 \hline
 Beale & 1000 & 0 & \textbf{0.0363} & 0.0635 & 1.0368 & 0.1174 \\
 \hline
 Cross in tray & 10000 & -2.0626 & \textbf{-2.0626} & \textbf{-2.0626} & -1.7594 & \textbf{-2.0626} \\
 \hline
Drop wave & 10000 & -1 & \textbf{-0.9841} & -0.9790 & -0.3781 & -0.9460 \\
\hline
Goldstein-Price & 1000 & 3 & \textbf{4.080} & 5.4463 & 70.427 & 6.240 \\
\hline
Griewank & 10000 & 0 & \textbf{0.1033} & 0.1471 & 21.920 & 1.0461 \\
\hline
Levy & 10000 & 0 & \textbf{0.1749} & 2.0717 & 20.081 & 1.5805 \\
\hline
Michealwicz & 10000 & -4.687 & -4.5119 & -4.0539 & -3.0879 & \textbf{-4.5187}\\
\hline
Rastrigin & 10000 & 0 & 12.458 & 14.298 & 55.091 &  \textbf{11.770} \\
\hline
Rosenbrock & 10000 & 0 & 5.3834 & 7776.2 & 12.612 & \textbf{1.1960} \\
\hline
Schwefel & 10000 & 0 & 1002.1 & 1083 & 1621.5 & \textbf{686.57} \\
\hline
Sphere & 1000 & 0 & \textbf{0} & \textbf{0} & \textbf{0} & \textbf{0} \\
\hline
\end{tabular}
\caption{Results of the optimization using different optimizers on different test functions. As one can see, the HOPSO method outperforms the standard PSO methods in all cases and in most of the cases its results are the best among all tested methods.}
\label{tab: Table2}
\end{table*}

\section*{Acknowledgment}
The authors would like to thank Martin Friak and his team for valuable discussions. 

\bibliographystyle{unsrt}
\bibliography{refs}

\begin{IEEEbiography}[{\includegraphics[width=1in,height=1.25in,clip,keepaspectratio]{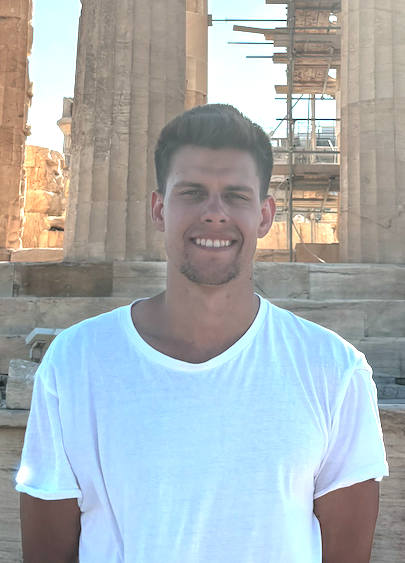}}]{Yury Chernyak} received a B.A. degree in Mathematics and Physics at Hartwick College, and his M.S. degree in Physics from SUNY University at Albany in 2022. He began his doctoral studies at the Institute of
Physics, Slovak Academy of Sciences in 2023.
\end{IEEEbiography}

\begin{IEEEbiography}[{\includegraphics[width=1in,height=1.25in,clip,keepaspectratio]{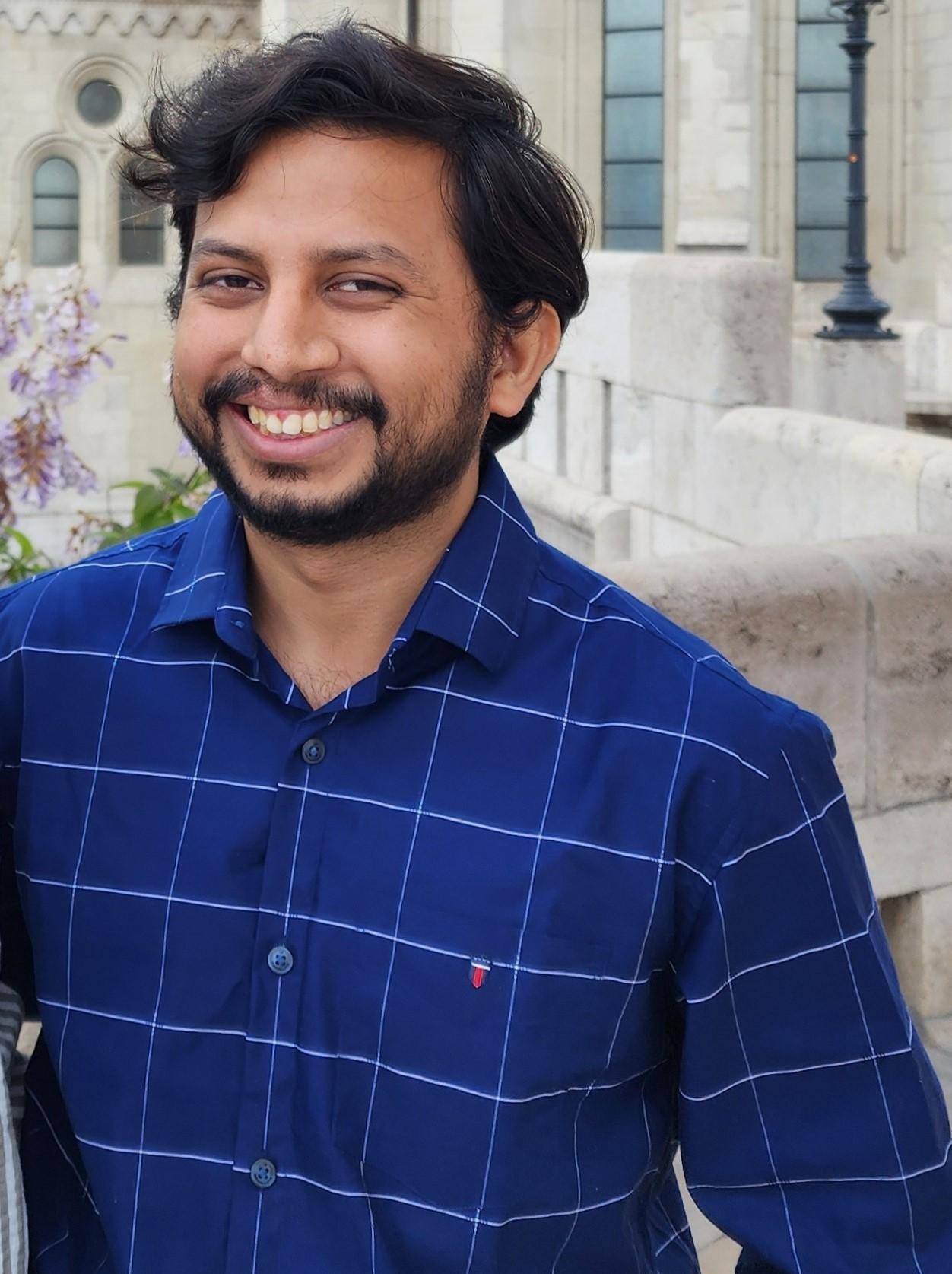}}]{Ijaz Ahamed Mohammad} 
is a senior Physics PhD student specializing in quantum computation. He finished his B.S and M.S in Physics at Indian Institute of Science Educational Research(IISER), Mohali, India in 2021. He was an INSPIRE-SHE scholarship recipient from 2016-2018. He started his doctoral studies at Institute of Physics, Slovak Academy of Sciences in 2021. He has contributed two research articles in the field of quantum information and computation.

\end{IEEEbiography}

\begin{IEEEbiography}
[{\includegraphics[width=1in,height=1.25in,clip,keepaspectratio]{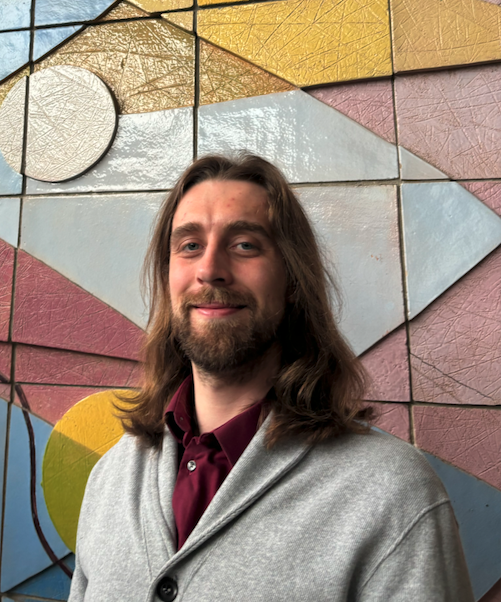}}]{Nikolas Masnicak} Recieved his B.S. and M.S. degree in theoretical physics from Masaryk University in Brno, Czech Republic in 2022. He started his doctoral studies at Institute of Physics, Slovak Academy of Sciences in 2023.
\end{IEEEbiography}

\begin{IEEEbiography}[{\includegraphics[width=1in,height=1.25in,clip,keepaspectratio]{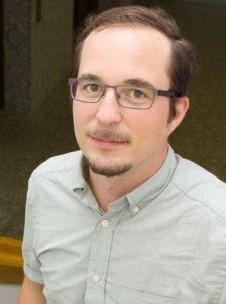}}]{Matej Pivoluska} is a Senior Engineer at Quantum Technology Laboratories in Vienna, specializing in quantum optics, quantum cryptography, quantum information processing, and quantum computing.
With a PhD in Informatics from Masaryk University, he has held various research positions, including at the Austrian Academy of Sciences and the Slovak Academy of Sciences.
Pivoluska has contributed extensively to high-dimensional quantum key distribution, quantum entanglement, and quantum computing, with over 30 published articles.

\end{IEEEbiography}

\begin{IEEEbiography}[{\includegraphics[width=1in,height=1.25in,keepaspectratio]{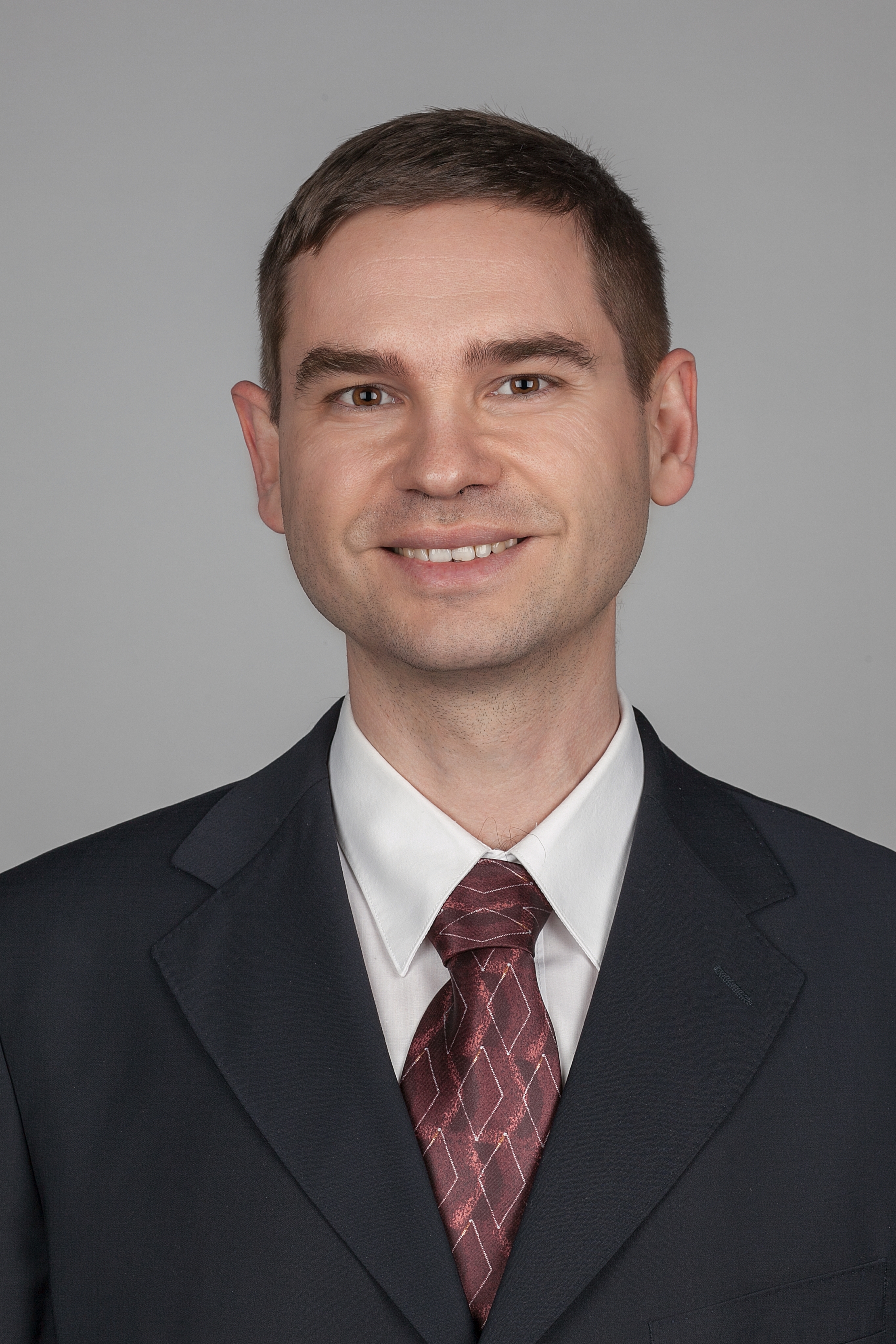}}]{Martin Plesch} is a physicist specializing in complex physical systems and quantum information theory. He is an independent researcher at the Institute of Physics Slovak Academy of Sciences and a Professor at Matej Bel University in Bratislava. With a PhD from the Institute of Physics Slovak Academy of Sciences, Prof. Plesch has held significant roles, including Head of the Department of Complex Physical Systems and Marie Curie Fellow at Masaryk University Brno. He has received numerous accolades for his research and educational contributions, including the Prize for Popularization of Science and the ``Social Innovator” award. Prof. Plesch is also actively involved in international scientific committees and educational initiatives, serving as a President of the International Young Physicists' Tournament and the World Federation of Physics.

\end{IEEEbiography}

\EOD

\end{document}